\documentclass[11pt]{article}

\usepackage[]{EMNLP2023}

\usepackage{times}
\usepackage{latexsym}
\usepackage[accsupp]{axessibility}

\usepackage[T1]{fontenc}

\usepackage[utf8]{inputenc}

\usepackage{microtype}

\usepackage{inconsolata}

\usepackage{svg}
\usepackage{amsmath}
\usepackage{comment}
\usepackage{graphicx}
\usepackage{babel,blindtext,multicol}
\usepackage{subcaption}
\usepackage{cleveref}
\usepackage{booktabs}
\usepackage{adjustbox}
\usepackage{lipsum,graphicx,subcaption}

\newcommand{\isidora}[1]{{\color{red}{\small\bf\sf [Isidora: #1]}}}

\title{Direct Neural Machine Translation with Task-level Mixture of Experts models}

\author{Isidora Chara Tourni \thanks{Work completed while interning at Google.} \\
  Boston University \\
  \texttt{isidora@bu.edu} \\\And
  Subhajit Naskar \\
  Google\\
  \texttt{snaskar@google.com} \\}

\begin{document}
\maketitle
\begin{abstract}
Direct neural machine translation (direct NMT) is a type of NMT system that translates text between two non-English languages. Direct NMT systems often face limitations due to the scarcity of parallel data between non-English language pairs. Several approaches have been proposed to address this limitation, such as multilingual NMT and pivot NMT (translation between two languages via English).
Task-level Mixture of expert models (Task-level MoE), an inference-efficient variation of Transformer-based models, has shown promising NMT performance for a large number of language pairs. In Task-level MoE, different language groups can use different routing strategies to optimize cross-lingual learning and inference speed. In this work, we examine Task-level MoE's applicability in direct NMT and propose a series of high-performing training and evaluation configurations, through which Task-level MoE-based direct NMT systems outperform bilingual and pivot-based models for a large number of low and high-resource direct pairs, and translation directions. Our Task-level MoE model with 16 experts outperforms bilingual NMT, Pivot NMT models for 7 language pairs, while pivot-based models still performed better in 9 pairs and directions.
\end{abstract}

\section{Introduction}



\begin{figure*}[tp]
\centering
\includegraphics[scale=0.5]{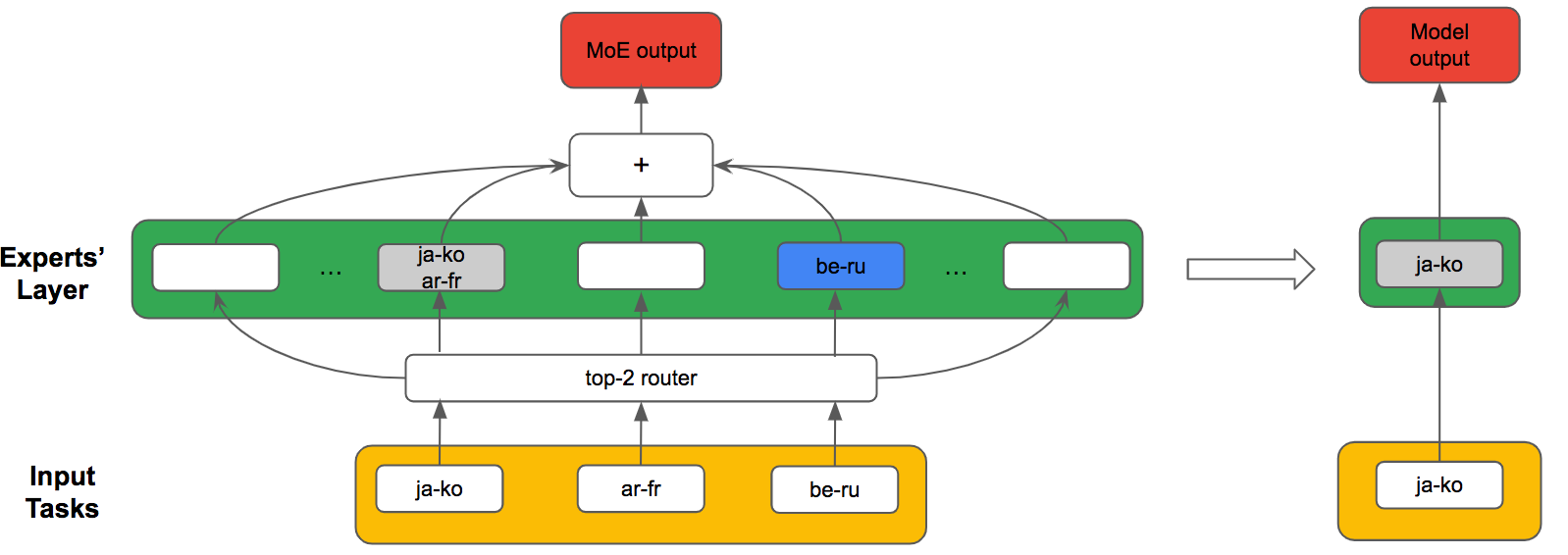}
\caption{Task-level MoE model, with LP - based routing; each Language Pair is routed through a top-2 router to an expert in the model experts' layer. From a pretrained Task-level MoE model we can extract a smaller dense network specializing in a certain task, e.g. ja-ko NMT.}
\label{fig:taskmoes}
\end{figure*}

NMT \citep{sutskever2014sequence,bahdanau2014neural,firat2016multi,johnson2017google,zoph-knight-2016-multi,bahdanau2014neural} has made remarkable progress in recent years, particularly with the advent of the Transformer architecture \citep{vaswani2017attention}, which has shown impressive results in bilingual and multilingual translation tasks \citep{aharoni-etal-2019-massively, hassan2018achieving,tang-etal-2021-multilingual}. Large Multilingual NMT models in particular have shown noteworthy performance and can serve translations between any languages, even for language pairs not seen in training, for which though NMT quality is often still low. The focus so far has primarily been on English-centric NMT, as there is an abundance of English-centric MT data \citep{tiedemann2018emerging} but few parallel data for non-English (low-resource) language pairs. Hence the need for improving methods for Direct NMT, i.e. NMT between non-English languages is evident \citep{yang-etal-2021-improving-multilingual,zhang2020improving}. 

Direct NMT methods, when no xx--yy data is available (where xx represents the source language code, and yy represents the target language code),
can be categorized into Zero-shot and Zero-resource approaches. 
In Zero-shot NMT \citep{johnson2017google,gu2019improved,wang-etal-2021-rethinking-zero,philip-etal-2020-monolingual}, translation occurs as follows: either through a bridging language, typically English, where models are trained on En--yy and xx--En datasets without ever being exposed to xx--yy data, while evaluation takes place on the few existing xx--yy pairs; or evaluating large multilingual models on xx--yy pairs, with neither xx or yy seen during model training. Zero-resource NMT \citep{currey2019zero,firat2016zero,bapna2022building}, on the other hand, generates pseudo-parallel data between direct language pairs, utilizing these synthetic xx--yy datasets during training. Yet bilingual or pivot-based models for direct NMT, though showing impressive NMT quality, are not practical and cost-efficient, especially as the number of languages for translation increases. Additionally, relying on related languages when evaluating a multilingual model on an unseen pair does not guarantee good NMT performance.
At the same time, scaling up to multilingual models of billions of parameters, trained on very large datasets of multiple language pairs, induces large training and inference costs. 

Sparse models, and specifically, sparsely activated Mixture-of-Experts (MoE) models \citep{shazeer2017outrageously,lepikhin2020gshard,zoph2022designing,ryabinin2020crowdsourced}, shape a promising direction in creating a universal translation model between all languages, while addressing efficiency concerns. We are the first to investigate performance of a specific type of MoE models, Task-level MoE models \citep{kudugunta2021beyond} in the context of Direct NMT, focusing on the impact of different task-to-expert mapping methods on model performance and expert utilization.
In our experiments, we train large multilingual Task-level MoE models across various configurations, including Zero-shot, Zero-resource, and scenarios where direct language pair data is available.
We conduct a comparative analysis of BLEU scores across Task-level MoE models with different mapping strategies and expert counts, as well as bilingual NMT and pivot-based NMT models; our findings serve as a valuable resource for determining the optimal model configuration to achieve high-quality translation results in various direct language pairs. 
By training large sparse multilingual Task-level MoE models which perform well for direct pairs, and showcase training efficiency, we may be able to extract smaller expert-specific dense models, which specialize in translating between Language Pairs or translating into one particular language. These models are ready for deployment and can replace bilingual or pivot-based models (Figure \ref{fig:taskmoes}). 
Prior work on Task-level MoEs \cite{kudugunta2021beyond} demonstrates the inference efficiency of task-level routing, compared to other approaches (token- or sentence-level), for the task of Multilingual NMT between English-centric pairs (En-yy and xx--En). To the best of our knowledge, we're the first ones to examine performance of Task-level MoE models in Direct NMT, offering a practical solution for expanding the coverage and quality of NMT across diverse language pairs.

\section{Related Work}


\subsection*{Direct Neural Machine Translation}


\citet{firat2016zero} first proposed a multilingual multi-way zero-resource NMT model which outperformed previous pivot-based approaches, laying the foundation for numerous subsequent zero-shot and/or zero-resource NMT methods. Various researchers have since expanded upon this foundational work, exploring different techniques and strategies for improving direct NMT \citep{ji2020cross,johnson2017google,zhang2020improving,chen2017teacher,lu2018neural,al2019consistency,rios2020subword,zheng2017maximum,gu2019improved,arivazhagan2019missing,kim2019pivot,cheng2019joint,chen2018zero,currey2019zero,ballesteros2003addressing,lakew2021zero}.
More specifically, \citet{freitag2020complete} constructed a multi-way aligned training set from existing training corpora, aligning examples in different language pairs with identical source or target sentences. They introduced a hierarchical data sampling strategy to prevent the over-representation of English in the training set and employed a temperature-based strategy to select target languages, while choosing source languages uniformly. Their complete Multilingual NMT  models demonstrated improved performance on direct pairs and comparable to baseline results on English-centric pairs.
\citet{liu2020improving} discretized the encoder output latent space into entries in a learned codebook, enabling the representation of source sentences in a common language and increasing model robustness and performance in zero-shot setups.
More recently, \citet{eriguchi2022building} 
pre-trained and fine-tuned a many-to-many multilingual NMT model that surpassed several pivoting and bilingual baselines for Direct NMT, and \citet{elnokrashy2022language}, presented a novel approach in which source and target language tokens are appended to the encoder input, while target language tokens are added to the decoder input. This differs from previous methods that used Source-only tokens and beginning of Sentence tokens for the encoder and decoder inputs, respectively, and aims to improve performance of Direct NMT models by providing more relevant context to both the encoder and decoder.
 
Latest works include the approach of \citet{Yang_2022}, who unified three distinct models to guide the student model during Direct NMT, aiming to improve its performance by harnessing the strengths of multiple models; 
and \newcite{xu2022eag}, who proposed a Direct NMT method comprising two stages: extraction of highly similar examples and their translations across different languages to construct a multi-way aligned parallel dataset, which enables the model to learn from a richer set of translation examples, and generation of additional aligned examples using a well-trained generative model. 

\begin{table*}[!htbp]
\centering
\small 
\adjustbox{max width=\textwidth}{
\begin{tabular}{l ll|ll|ll|ll}
\toprule
&\multicolumn{1}{c}\textbf{ja--th}
&&\multicolumn{1}{c}\textbf{bg--mk}
&&\multicolumn{1}{c}\textbf{ru--tr}
&&\multicolumn{1}{c}\textbf{fr--ar}

\\
\cmidrule{2-3}\cmidrule{4-5}\cmidrule{6-7}\cmidrule{8-9}

&$\leftarrow$&$\rightarrow$&   
 $\leftarrow$&$\rightarrow$&$\leftarrow$&$\rightarrow$&$\leftarrow$&$\rightarrow$
\\ \hline
Task-level MoE 
&\textbf{21.51}
&25.36
&26.44
&23.29
&10.24
&8.72
&19.37
&\textbf{15.06}
\\ 

Bilingual 
&6.24
&7.26
&24.47
&21.28
&9.53
&7.58
&19.6
&9.71
\\  
Pivot-level
&18.34
&\textbf{39.28}
&\textbf{28.24}
&\textbf{27.03}
&\textbf{18.44}
&\textbf{13.84}
&\textbf{21.57}
&14.91
\\
\toprule

Number of Experts
&16
&64
&16
&16
&16
&16
&16
&16
\\

Decoding mapping method
&lp\_b
&tl\_b
&tl\_a
&tl\_a
&tl\_a
&tl\_a
&tl\_a
&tl\_a
\\
\bottomrule
\end{tabular}}\\
\adjustbox{max width=0.9\textwidth}{
\begin{tabular}{l ll|ll|ll|ll}
\toprule
&\multicolumn{1}{c}\textbf{zh--ko}
&&\multicolumn{1}{c}\textbf{ja--vi}
&&\multicolumn{1}{c}\textbf{ja--ko}
&&\multicolumn{1}{c}\textbf{ja--zh}
\\
\cmidrule{2-3}\cmidrule{4-5}\cmidrule{6-7}\cmidrule{8-9}


&$\leftarrow$&$\rightarrow$&$\leftarrow$&$\rightarrow$&$\leftarrow$&$\rightarrow$&$\leftarrow$&$\rightarrow$\\ \hline
Task-level MoE
&\textbf{34.35}
&28.70
&\textbf{25.17}
&19.18
&\textbf{42.64}
&\textbf{41.38}
&\textbf{34.94}
&22.52
\\ 
Bilingual 
&5.71
&3.16
&8.20
&19.6
&6.12
&4.47
&3.16
&7.26
\\  
Pivot-level
&28.13
&\textbf{35.24}
&18.97
&\textbf{19.61}
&21.49
&36.69
&19.98
&\textbf{27.46}
\\
\toprule
Number of Experts
&16
&16
&16
&64
&16
&16
&16
&64
\\

Decoding mapping method
&lp\_b
&lp\_b
&lp\_b
&lp\_a
&lp\_b
&lp\_b
&lp\_b
&lp\_a
\\
\bottomrule
\end{tabular}}

\caption{Best BLEU scores from our Task-level MoE models trained with 16 or 64 experts, for direct pairs. We compare to Bilingual and Pivot-level NMT models' scores and state the Task-level MoE model/setup under which best scores were obtained. overall best scores per pair are highlighted. lp\_x decoding mapping method implies a model trained with Language Pair (LP) - based task to expert mapping, and tl\_x implies target language (TL) - based task to expert mapping.}
\vspace{-2mm}
\label{table:bleu_table_best_LP_tl_dir}
\vspace{-0.5em}
\end{table*}

\begin{table*}[!htbp]
\small 
\centering
\adjustbox{max width=\textwidth}{
\begin{tabular}{ll ll|ll|ll|ll}
\toprule
&&\multicolumn{1}{c}\textbf{ja--th}
&&\multicolumn{1}{c}\textbf{bg--mk}
&&\multicolumn{1}{c}\textbf{ru--tr}
&&\multicolumn{1}{c}\textbf{fr--ar}
\\
\cmidrule{3-4} \cmidrule{5-6}\cmidrule{7-8}\cmidrule{9-10}
&
&$\leftarrow$&$\rightarrow$&$\leftarrow$&$\rightarrow$&$\leftarrow$&$\rightarrow$&$\leftarrow$&$\rightarrow$
\\ \hline

Task-level MoE (LP) 
&\textbf{lp\_a}
&0.47
&2.26
&3.21
&22.75
&1.55
&0.74
&-
&-
\\ 

&\textbf{lp\_b}
&\textbf{21.51}
&18.92
&3.07
&20.43
&4.86
&0.83
&19.37
&\textbf{15.06}
\\
&\textbf{lp\_c}
&0.42
&1.50
&5.97
&18.32
&5.57
&0.80
&2.40
&0.86

\\
Task-level MoE (TL) 
&\textbf{tl\_a}
&1.15
&0.78
&26.44
&23.29
&10.24
&8.72
&5.36
&1.18
\\
&\textbf{tl\_b}
&1.39
&6.09
&5.19
&10.15
&4.45
&0.86
&5.62
&0.81
\\
Bilingual 
&&6.24
&7.26
&24.47
&21.28
&9.53
&7.58
&19.60
&9.71
\\  
Pivot-level
&&18.34
&\textbf{39.28}
&\textbf{28.24}
&\textbf{27.03}
&\textbf{18.44}
&\textbf{13.84}
&\textbf{21.57}
&14.91
\\ \hline
\tiny{Number of train sentences}
&&\text{\tiny{6,200}}
&&\text{\tiny{205,651}}
&&\text{\tiny{1,467,160}}
&&\text{\tiny{17,163,359}}
\\ \bottomrule
\end{tabular}}

\adjustbox{max width=0.9\textwidth}{
\begin{tabular}{ll ll|ll|ll|ll}
\toprule
&&\multicolumn{1}{c}\textbf{zh--ko}
&&\multicolumn{1}{c}\textbf{ja--vi}
&&\multicolumn{1}{c}\textbf{ja--ko}
&&\multicolumn{1}{c}\textbf{ja--zh}
\\
\cline{3-4}\cline{5-6}\cline{7-8}\cline{9-10}
& &  $\leftarrow$&$\rightarrow$&$\leftarrow$&$\rightarrow$&$\leftarrow$&$\rightarrow$&$\leftarrow$&$\rightarrow$
\\ \hline
Task-level MoE (LP) 
&\textbf{lp\_a}
&2.81
&4.09
&0.77
&5.32
&3.33
&11.78
&1.67
&9.06
\\ 
&\textbf{lp\_b}
&\textbf{34.35}
&28.70
&\textbf{25.17}
&1.78
&\textbf{42.64}
&\textbf{41.38}
&\textbf{34.93}
&2.50
\\
&\textbf{lp\_c}
&2.35
&3.04
&0.57
&1.31
&2.71
&10.14
&1.30
&2.26
\\
Task-level MoE (TL) 
&\textbf{tl\_a}
&4.51
&0.61
&0.87
&14.31
&3.87
&1.30
&1.86
&18.77
\\ 
&\textbf{tl\_b}
&4.44
&10.7
&1.34
&4.80
&3.91
&19.91
&2.79
&4.06
\\
Bilingual 
&
&5.71
&3.16
&8.20
&19.6
&6.12
&4.47
&3.16
&7.26
\\  
Pivot-level
&
&28.13
&\textbf{35.24}
&18.97
&\textbf{19.61}
&21.49
&36.69
&19.98
&\textbf{27.46}
\\
\hline
\tiny{Number of train sentences}
&&\text{\tiny{498,968}}
&&\text{\tiny{604,940}}
&&\text{\tiny{974,896}}
&&\text{\tiny{1,339,622}}\\
\bottomrule
\end{tabular}}
\caption{BLEU scores of Task-level MoE models trained with 16 experts, for direct pairs, in both directions, for models with Language Pair (LP) - or target language (TL) - based routing during training, lp\_a, lp\_b, lp\_c and tl\_a, tl\_b mapping during inference, respectively. We compare to Bilingual and Pivot-level NMT models' scores and highlight the model/setup under which best scores were obtained.}
\vspace{-2mm}
\label{table:bleu_table_16_LP_tl_dir}
\vspace{-0.5em}
\end{table*}

\begin{table*}[!htbp]
\small 
\centering
\adjustbox{max width=\textwidth}{
\begin{tabular}{ll ll|l|ll|ll|l|ll|l}
\toprule
&&\multicolumn{1}{c}\textbf{ja-th}
&&bg-mk
&\multicolumn{1}{c}\textbf{fr-ar}
&&\multicolumn{1}{c}\textbf{zh--ko}
&& ja--vi
&\multicolumn{1}{c}\textbf{ja--ko}
&& ja--zh
\\
\cmidrule{3-3}\cmidrule{4-4}\cmidrule{5-6}\cmidrule{7-8}\cmidrule{9-9}\cmidrule{10-11}\cmidrule{12-13}
&&\centering 

$\leftarrow$&$\rightarrow$&$\rightarrow$&$\leftarrow$&$\rightarrow$&$\leftarrow$&$\rightarrow$&$\leftarrow$ &$\leftarrow$&$\rightarrow$&$\leftarrow$
\\ \hline
Task-level MoE (LP) 
&\textbf{lp\_a}
&16.33
&17.50
&19.18
&-
&-
&21.34
&19.08
&\textbf{19.18}
&\textbf{25.3}
&25.25
&\textbf{22.52}
\\ 
&\textbf{lp\_b}
&0.44
&2.79
&19.07
&4.47
&2.13
&2.05
&2.51
&0.47
&0.96
&4.46
&0.71
\\
&\textbf{lp\_c}
&0.36
&0.50
&19.04
&2.69
&1.18
&0.59
&0.47
&0.45
&0.80
&2.05
&0.56
\\
Task-level MoE (TL) 
&\textbf{tl\_a}
&8.30
&25.33
&20.29
&7.37
&5.38
&12.61
&20.27
&5.95
&14.25
&26.39
&0.01
\\
&\textbf{tl\_b}
&8.33
&25.36
&20.35
&7.32
&5.50
&12.43
&20.34
&5.96
&14.26
&26.40
&10.49
\\
Bilingual 
&
&6.24
&7.26
&21.28
&9.71
&19.60
&5.71
&3.16
&8.20
&6.12
&4.47
&3.16
\\  
Pivot-level
&
&\textbf{18.34}
&\textbf{39.28}
&\textbf{27.03}
&\textbf{14.91}
&\textbf{21.57}
&\textbf{28.13}
&\textbf{35.24}
&18.97
&21.49
&\textbf{36.69}
&19.98
\\ \hline
\tiny{Number of train sentences}
&
&\text{\tiny{6,200}}
&\text{\tiny{205,651}}
&&\text{\tiny{17,163,359}}
&&\text{\tiny{498,968}}
&&\text{\tiny{604,940}}
&&\text{\tiny{974,896}}
&\text{\tiny{1,339,622}} \\
\bottomrule
\end{tabular}}
\caption{BLEU scores of Task-level MoE models trained with 64 experts, for direct pairs, in both directions, for models with Language Pair (LP) - or target language (TL) - based routing during training, lp\_a, lp\_b, lp\_c and tl\_a, tl\_b mapping during inference, respectively. We compare to Bilingual and Pivot-level NMT models' scores and highlight the model/setup under which best scores were obtained.}
\vspace{-2mm}
\label{table:bleu_table_64_LP_tl_dir}
\vspace{-0.5em}
\end{table*}

\subsection*{Conditional Computation, Sparsity and Mixture-of-Experts models}
\citet{fedus2022review} extensively review  sparse Mixture-of-Experts (MoE) models in Deep Learning, starting from the foundational work of \citet{shazeer2017outrageously}, \citet{lepikhin2020gshard} and
\citet{fedus2021switch}. Their work delves into the scaling characteristics, routing algorithms, and training improvements proposed for MoE models, as well as the latest applications of Sparse MoE models across various domains, such as Natural Language Processing (NLP) \citep{zoph2022designing,chowdhery2022palm,lee2022sparse,du2022glam}, Computer Vision \citep{riquelme2021scaling,DBLP:journals/corr/abs-2009-13239,wu2022residual,hwang2022tutel}, Speech Recognition \citep{DBLP:journals/corr/abs-2105-03036,you2022speechmoe2}, and Multimodal Learning \citep{https://doi.org/10.48550/arxiv.2206.02770}.
In NLP, numerous studies have explored the potential of Sparse MoE models for diverse applications. \citet{kudugunta2021beyond} introduced the concept of task-level routing in MoE models, achieving improved translation scores compared to large multilingual models with the same inference cost and dense student models distilled from token- or position-level MoE models.
By using task-level MoE models in our experiments, we investigate and optimize their performance in direct pairs' NMT.

\section{Experiments}

\subsection*{Model}
We train sparse encoder-decoder Task - level MoE models,
with dim = 1024, hidden dim = 4096, 8 heads, 3 layers. Following \citet{kudugunta-etal-2021-beyond-distillation},
we similarly replace the Transformer Feed Forward Network (FFN) with a set of identical FFN experts. We use a 32,000 token SentencePiece vocabulary \citep{kudo2018sentencepiece}, shared in both the encoder and decoder of the model. We experiment with the number of experts, selecting either 16 or 64 experts, which correspond to models with approximately 1 billion and 3.5 billion parameters, respectively.
We investigate the impact of different task id to expert mapping methods, setting the task id to be either the Target Language or the Language Pair for translation.
Following \citet{kudo2018sentencepiece,kudugunta2021beyond}, data sampling temperature is T=5 during training, and we train 8 models for 2 million steps.

The training process has a batch size of 256, a maximum sentence length of 128. 
By examining performance of these models under different configurations, we want to better understand the relationship between the number of experts, task id to experts' mappings, and translation quality, ultimately informing the optimal design choices for Task-level MoE models in various scenarios.

\subsection*{Languages and Training Datasets}
We train our models using in-house production-scale datasets, consisting of English-centric, and Direct pairs' parallel sentences, covering 108 languages, including English. Our training data includes a total of 107 English-centric pairs, and 53 direct pairs, with data sizes for each language pair ranging from a few 
thousand to a few billion sentences for each pair. The number of sentences in each parallel pair in the train set is provided in Table \ref{table:train_num_of_sents} in the Appendix.

\subsection*{Evaluation Sets and Task Mapping}
For evaluation, we use internal test sets from two distinct sources: the Web Domain and Wikipedia. These test sets vary in size, containing anywhere from 500 to 5,000 sentences for each language pair.
In both training and evaluation datasets, we follow a specific preprocessing convention, prepending the source sentence with  <4xx> <2yy> tokens. These tokens serve as a cue for the model, helping it identify the source and target languages in context, and the model to better learn and adapt to various language pair combinations.

Our models use either the exact train set Language Pair (LP) or the target language (TL) of the pair as the task id for mapping translation pairs to different experts. 
In the former case, we directly map the specific language pair (En--yy, xx--En or xx--yy) to a task id, and have a total of 214 tasks (107 x 2 English-centric pairs) when training with English-only datasets, or 267 tasks (+ 53 direct Language Pairs) when incorporating Direct LPs.
On the other hand, when using the target language as the task id, we have 108 tasks (107 languages and English) in all scenarios. There we leverage and try to benefit from the shared characteristics of translations into the same target language.
Since we evaluate our models on direct (non-English) data Language Pairs, it is often the case that the exact mapping of the pair to an expert was not defined during training. To address this issue, we experiment with several approaches for mapping language pair xx--yy to a task id at inference time - in each method mentioned below we name the component used as task id:
\begin{itemize}
\item Language Pair (LP) as task id. 
\begin{itemize}
\item \textbf{lp\_mapping\_a (lp\_a)}: exact LP
\item \textbf{lp\_mapping\_b (lp\_b)}: {En--yy} LP
\item \textbf{lp\_mapping\_c (lp\_c)}: {xx--En} LP
\end{itemize}
\item Target language as task id
\begin{itemize}
\item \textbf{tl\_mapping\_a (tl\_a)}: yy
\item \textbf{tl\_mapping\_b (tl\_b)}: xx
\end{itemize}
\end{itemize}

\section{Results \& Discussion}
\begin{figure*}[tp]
\tiny
\centering
\includegraphics[scale=0.5]{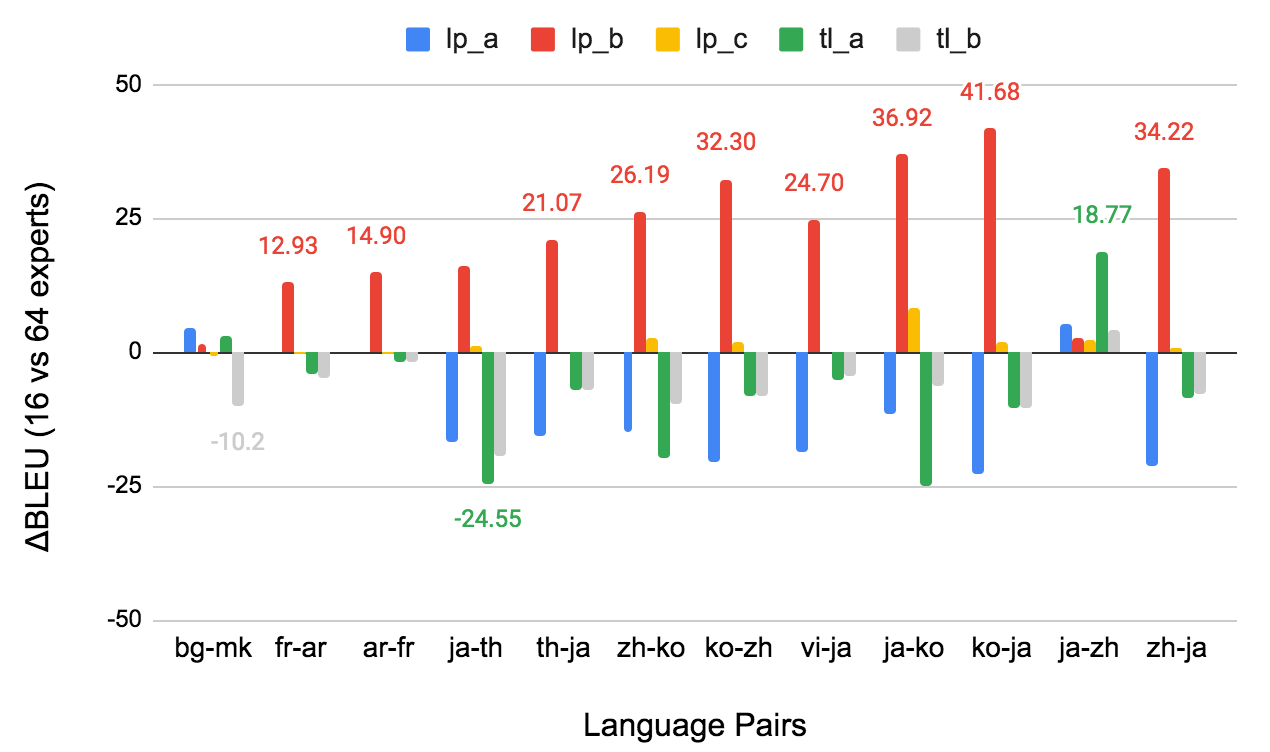}
\caption{Difference between BLEU scores of Task-level MoE models trained with 16 and 64 experts, for direct pairs, for models with Language Pair (LP) - based routing during training, lp\_a, lp\_b, lp\_c mapping during inference, respectively, and models with Target Language (TL) - based routing during training, tl\_a, tl\_b mapping during inference. For each pair, we mark the BLEU score value of the model which shows the largest difference among the 16 and 64 experts Task-level MoE variation.}
\label{fig:bleu_diff_16_64}
\tiny
\end{figure*}

In this study, we have two primary baselines for comparison: bilingual NMT models and pivot-based NMT models, which are trained using a bridging strategy through English. We use BLEU~\citep{papineni-etal-2002-bleu} for evaluating the translation quality of all direct language pairs. We are aware that learned metrics like COMET~\cite{rei-etal-2022-comet} and BLEURT~\cite{sellam-etal-2020-bleurt} show higher correlation with human judgements for high resource, English centric language pairs~\cite{freitag-etal-2022-results}. However, it is unknown how well learned metrics perform on low resource languages and thus we decide to stick with BLEU for the main part of the paper.
Tables \ref{table:bleu_table_16_LP_tl_dir} and \ref{table:bleu_table_64_LP_tl_dir} summarize results for Task-level MoE models with either 16 or 64 experts.

In Table \ref{table:bleu_table_best_LP_tl_dir}, we show the specific configurations that led to the highest BLEU scores for Task-level MoE models - comparing those to the performance of bilingual and pivot-based models, we aim to emphasize the best BLEU scores for each language pair. This analysis allows us to identify the most effective models for each pair and gain insights into how different model configurations and expert counts contribute to translation performance, so we can better understand the advantages and limitations of Task-level MoE models with varying numbers of experts, as well as bilingual and pivot-based NMT models.
Our results ultimately help inform the choice of translation models and strategies to optimize performance for different language pairs

Performance of NMT models varies significantly depending on the pair, and on the direction of translation - certain pairs show large discrepancies between translation directions.
We see that pivot-based models perform better than bilingual ones for the majority of pairs. Bilingual NMT and pivot-based NMT models demonstrate strong results in several instances, while Task-level MoE models show mixed performance across different Language Pairs.

\subsection*{Which model performs best for each pair?}

Table \ref{table:bleu_table_best_LP_tl_dir} results reveal that for the majority of direct language pairs, the highest Task-level MoE BLEU scores are achieved using a model with 16 experts, with the tl\_a and lp\_b mapping-to-experts methods surpassing other approaches. It is particularly noteworthy that for pairs of languages belonging to closely related language families (Japanese, Chinese, Korean, Vietnamese, and Thai), the LP-mapping-based models seem to perform best. In contrast, TL-based models tend to outperform the rest of the other language pairs.
In this case, Task-level MoE models show superior performance in the forward translation direction across all pairs. Meanwhile, pivot-based models yield the best results in the backward translation direction for four out of five pairs in those related languages. This observation suggests that different mapping-to-experts strategies may be more effective for certain language pairs or translation directions, depending on the linguistic relationships and unique characteristics of the languages involved; a tailored approach in employing Task-level MoE models and mapping-to-experts methods is the key to optimizing translation performance for different language pairs and directions.

\subsection*{Comparison of different task to expert mapping methods}
From Table \ref{table:bleu_table_16_LP_tl_dir}, pivot-based models, Task-level MoE (LP) models with lp\_b task id mapping and Task-level MoE (TL) models with tl\_a task id mapping (both trained with 16 experts) are primarily the best performing models among all pairs.

Task-level MoE models with LP task id to expert routing and lp\_b mapping at inference time excel in translations to Japanese, from Thai, Vietnamese, Korean, and Chinese (+3.17, +6.2, +21.15, +14.95 BLEU points over the pivot-level NMT model, respectively, and +15.27, +16.97, +36.52, +31.77 BLEU gain against bilingual models) as well as Korean-Chinese (ko-zh) (+6.22 BLEU compared to pivot-based model, +28.64 BLEU over the bilingual baseline) and French-Arabic (fr--ar) (+0.15 and +5.35 BLEU over the pivot and bilingual NMT models, respectively). This indicates the lp\_b mapping strategy is particularly effective for these language pairs; routing to the expert best at En--yy translation, where yy is each pair's target language, instead of routing to the expert explicitly trained on the exact direct pair, xx--yy, likely captures better the linguistic nuances of the pair and leads to improved translation scores. 

Task-level MoE models trained with the TL task-to-expert mapping approach and the tl\_a task id mapping, which leverages target language pairs for routing during inference, achieve the highest scores for 
translations involving Kazakh (kk), such as 
Kazakh-Russian (kk--ru) in both directions (+13.62 and +13.68 BLEU gain against pivot-based model, +0.34 and +12.34 BLEU points over bilingual models). Success of tl\_a mapping strategy suggests the model shows superior performance in those cases and when the pair is properly routed to the expert specialised in the pair's target language.

Table \ref{table:bleu_table_64_LP_tl_dir} shows results from our evaluation of 64 experts' Task-level MoE models. Interestingly, pivot-based methods stand out as top performing in the majority of pairs and directions. These pairs include Bulgarian--Macedonian (bg--mk), French--Arabic (fr--ar), and Japanese--Thai (ja--th), as well as Chinese-Korean (zh-ko) in both directions, and Japanese-Korean (ja--ko).
In all other pairs and directions, specifically when translating to Japanese from Vietnamese, Korean, and Chinese, Task-level MoE (LP) models with the lp\_a mapping approach emerge as the best-performing models (+0.21, +3.81, +2.54 BLEU points against the pivot-based model, respectively, and +10, +19.18, +19.36 BLEU gain over the bilingual baseline).

For certain Language Pairs with smaller amounts of training data 
bilingual NMT models achieve reasonable performance even with limited data and often outperform others.
On the other hand, we see that pivot-based NMT models perform well when there is a substantial amount of training data (e.g., Belarusian--Macedonian, and Russian--Turkish in both directions, Arabic--French, Chinese--Korean, Japanese--\{Thai, Vietnamese, Chinese\}). This might be due to the fact that these models leverage English as an intermediate bridging language between source and target languages, which is helpful when there's enough data to help in learning meaningful representations.



\subsubsection*{Comparison of models with different number of experts}
In Figure \ref{fig:bleu_diff_16_64}, we present the difference in BLEU scores on select direct language pairs for Task-level MoE models trained using Language Pair (LP) or Target Language (TL) based mapping, with 16 and 64 experts. We seek to visually emphasize differences in performance across models that utilize varying numbers of experts, in order to get insights into their relative effectiveness and understand how the choice of mapping strategy (LP or TL) and the number of experts (16 or 64) impacts translation performance. For the majority of pairs (Vietnamese--Japanese, and both directions of French--Arabic, Japanese--Thai, Chinese--Korean, Japanese--Korean, Japanese--Chinese), Task-level MoE models with 16 experts outperform those with 64. For those language pairs, with the exception of Japanese--Chinese, for which TL-based Task-level MoE with tl\_a shows best performance, LP-based Task-level MoE with lp\_b mapping show the largest gain when trained with fewer experts, corroborating our previous findings. For other pairs, such as Japanese--Thai, Japanese--Chinese, 
and Bengali--Macedonian, different model variations show a largest improvement when trained with either 16 or 64 experts. We also notice a significant, yet not maximum, BLEU score gain for a large number of pairs with the TL-based Task-level MoE model and the LP-based Task-level MoE model, when using tl\_a and lp\_a on the task-to-expert mapping during inference, respectively, between 16 and 64 experts' models.

\subsection*{Routing decisions analysis}

\begin{figure*}
\tiny
\centering
\vspace{-2mm}
\includegraphics[scale=0.1]{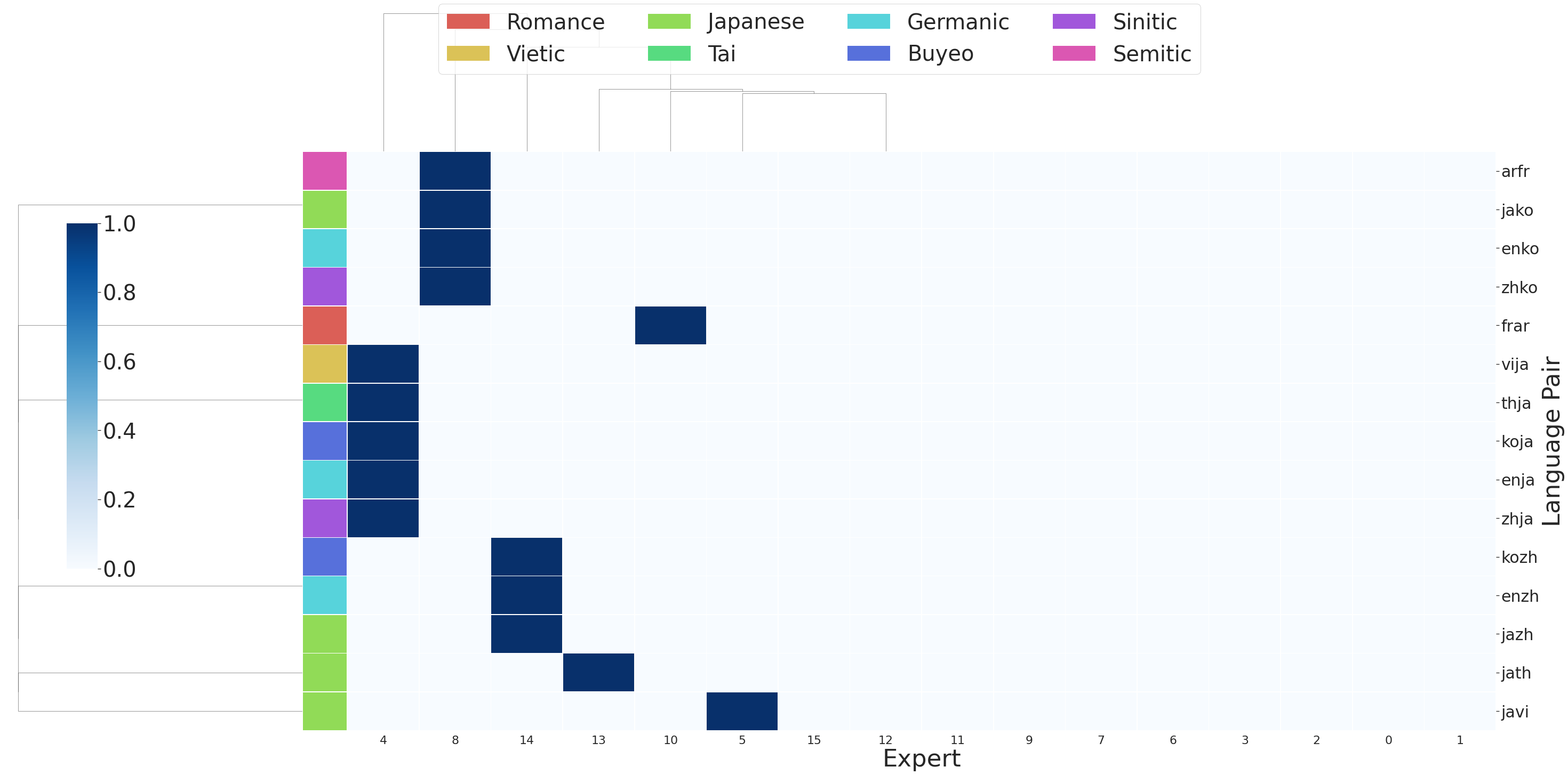} 
\vspace{-1.7mm}
\caption{Routing decisions of the last encoder layer of our Task-level MoE model with 16 experts, trained with pair target language to task id mapping, with tl\_a used during inference, for 1M steps.}
\label{fig:fig_16_tl_enc3_1m}
\end{figure*}

\begin{figure*}
\centering
\vspace{-2mm}
\includegraphics[scale=0.1]{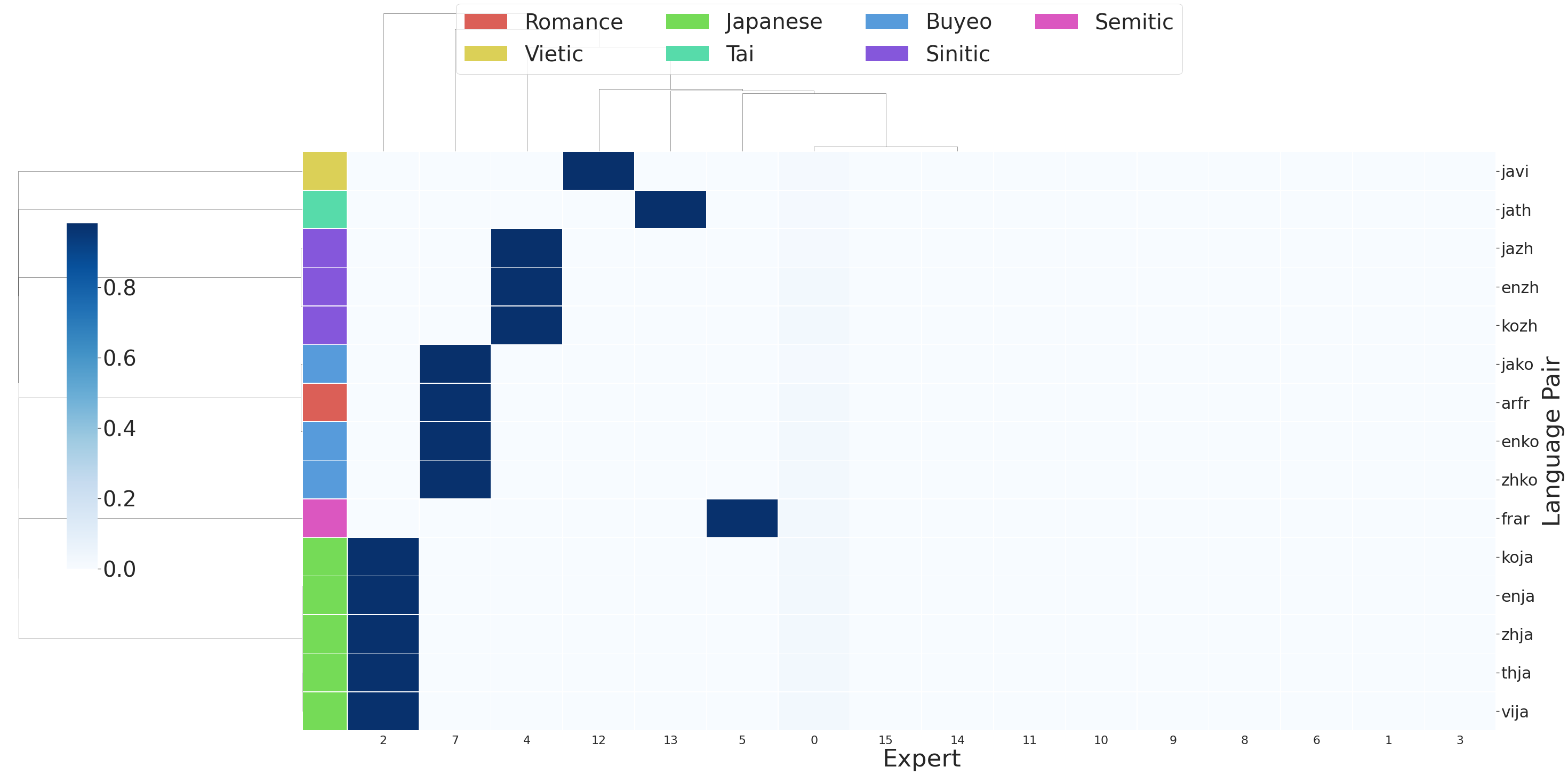}
\vspace{-1.7mm}
\caption{Routing decisions of the last decoder layer of our Task-level MoE model with 16 experts, trained with pair target language to task id mapping, with tl\_a used during inference, for 1M steps.}
\label{fig:fig_16_tl_dec3_1m}
\end{figure*}

\begin{figure*}
\centering
\vspace{-2mm}
\includegraphics[scale=0.1]{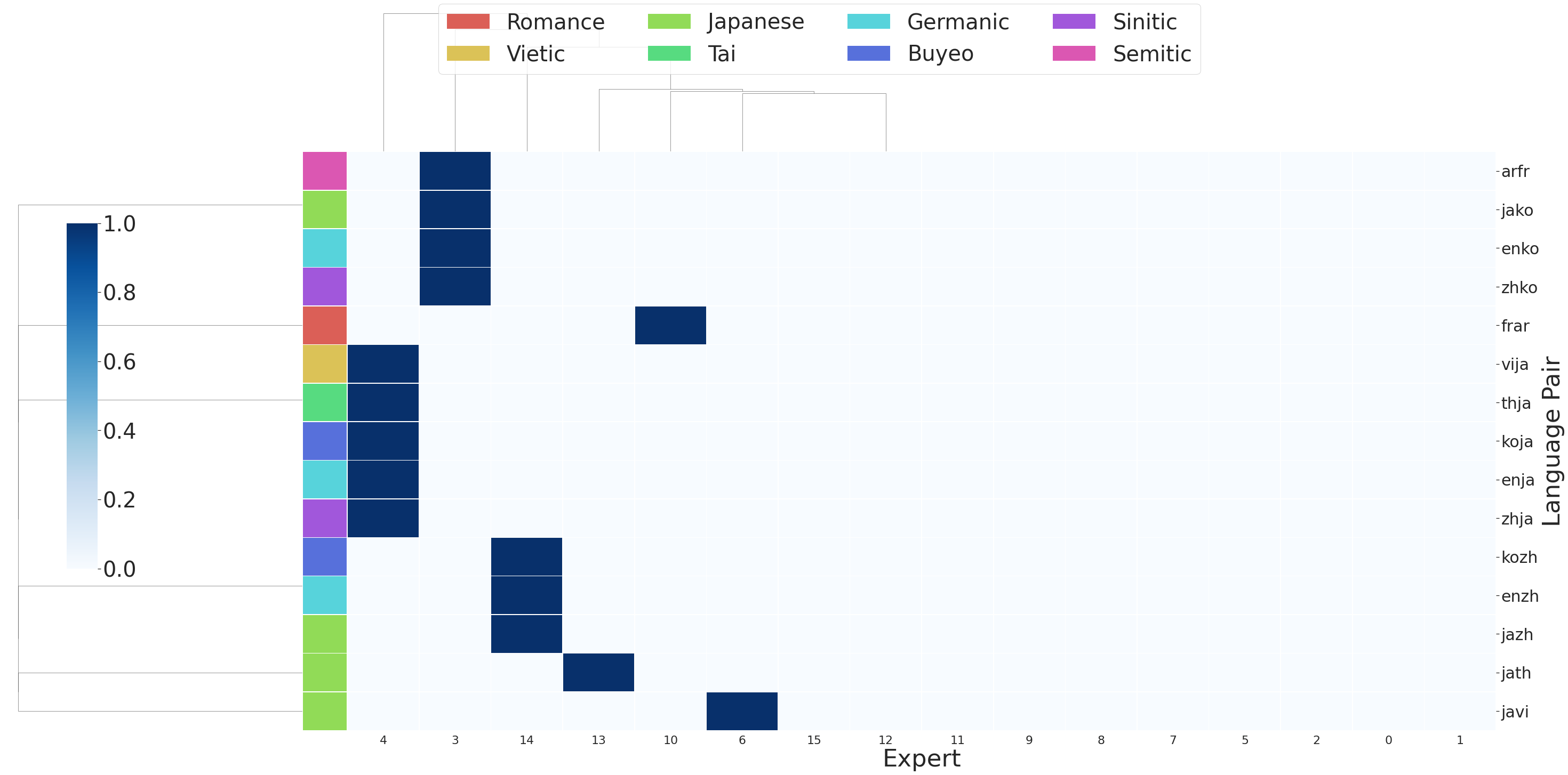}
\vspace{-1.7mm}
\caption{Routing decisions of the last encoder layer of our Task-level MoE model with 16 experts, trained with pair target language to task id mapping, with tl\_a used during inference, for 2M steps.}
\label{fig:fig_16_tl_enc3_2m}
\end{figure*}

\begin{figure*}
\centering
\vspace{-2mm}
\includegraphics[scale=0.1]{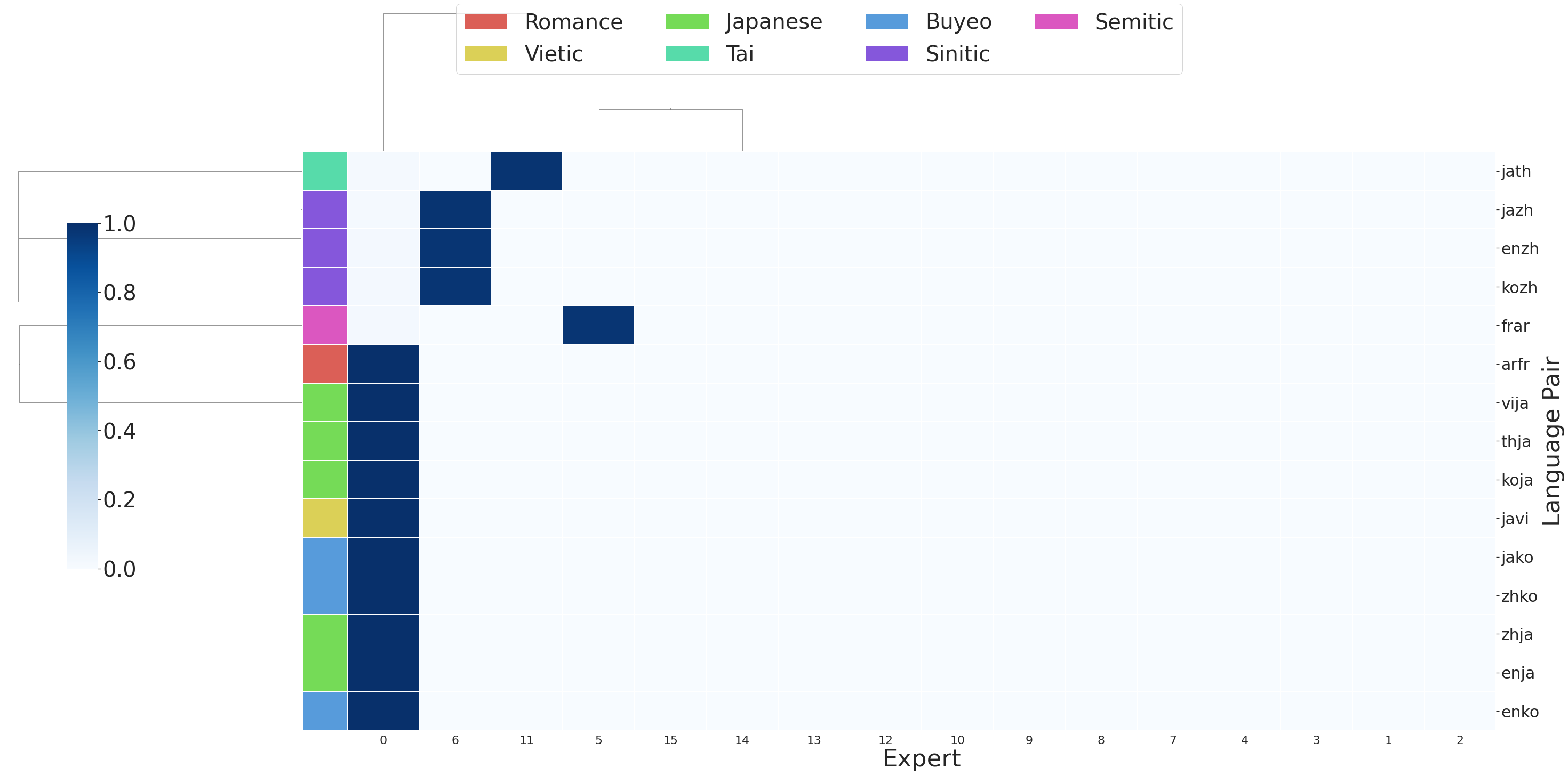}
\vspace{-1.7mm}
\caption{Routing decisions of the last decoder layer of our Task-level MoE model with 16 experts, trained with pair target language to task id mapping, with tl\_a used during inference, for 2M steps.}
\label{fig:fig_16_tl_dec3_2m}
\end{figure*}

In Figures \ref{fig:fig_16_tl_enc3_1m}, \ref{fig:fig_16_tl_dec3_1m},
\ref{fig:fig_16_tl_enc3_2m}, \ref{fig:fig_16_tl_dec3_2m} we show the experts' utilization for the last model's encoder and decoder layers for different pairs, for our TL-based Task-level MoE model, with tl\_a used as a task to expert mapping approach during training; The darker a cell is in each of the Figures, the more that expert is used by the specific language pair. We can observe that language pairs with the same target language get properly routed to the same expert, as expected. It's interesting to see that there are no major differences between the encoder's routing decisions across checkpoints, yet we notice changes in experts' assignment in the decoder during and at the end of training; at first, pairs with the same target language are mapped to the same different experts, yet as the model converges, we see a significant overlap in the selected experts, for pairs with different target languages. We also see a great overlap in experts chosen in the encoder during and at the end of training - the majority of experts are the same throughout model training and certain experts are preferred over others. This preference differs from the experts chosen in the decoder, where we see almost no overlap. It is also worth noting that the number of common experts between the encoder and the decoder is minimal, both during training and upon model convergence.


In the Appendix, we additionally show experts' utilization in the encoder and decoder of our 64 experts' TL-based Task-level MoE model. 

\section{Conclusions}
We conducted a thorough analysis of the usage of Task-level Mixture of Expert models for Direct NMT.
Our experiments reveal the strengths and weaknesses of different approaches and shed light on which configurations work best for specific direct language pairs. Our comparisons help in identifying best-performing models and offer valuable insights into how varying numbers of experts, and different task-to-expert mapping methods, during training and inference, can influence direct pairs' translation quality in Task-level MoE models.
Specifically, we noticed that Task-level MoE NMT models, along with pivot-based approaches, are frequently top performers for numerous direct language pairs. However, their performance varies between different pairs and translation directions. Besides observing NMT quality results, our analysis of expert’s usage throughout training also serves as an informative way to visualize and understand mapping of language pairs to model experts in the encoder and decoder throughout training. Future work can focus on enhancing this model for broader language coverage, including other low-resource languages, to further improve translation quality and efficiency.

\newpage \newpage \newpage
\section*{Limitations}
Training and evaluation of Task-level MoE models can be very challenging due to the models' size and complexity. There is a large number of parameters that require tuning, such as the number and size of experts, the number and diversity of languages, and of English-centric and direct Language Pairs in training, the batch size, the vocabulary size and the maximum sentence length. At the same time, it is time and computationally expensive to evaluate on a large number of direct pairs, train separate Bilingual and Pivot-level models for all to use as baselines, and also perform the experts' analysis and visualizations for those pairs. This automatically restricts the amount and breadth of results, and calls for further exploration of the model capabilities in the future.

\section*{Ethics Statement}
Working with large language models raises several ethical concerns in NLP research, particularly related to quality, bias and toxicity of the output result \cite{10.1145/3442188.3445922,chowdhery2022palm,blodgett-etal-2020-language,brown2020language}.
In the context of NMT, involvement of many stakeholders in the task, as well as the dangers arising from mistranslation of the original text need to be taken into consideration, since they can potentially affect the perception of the work and harm the interests of the different parts \citep{taivalkoski2019ethical,gambier2016border}.  Undoubtedly, the benefits of responsibly developed and deployed NMT systems lie in making an author’s work more accessible, enabling the transfer of ideas to other audiences, and enhancing both the reader’s capabilities and the translator’s role and sense of contribution \citep{besacier-2014-machine}, which are directions any work in the area should aim to focus on.

\newpage
\bibliography{custom}

\begin{thebibliography}{63}
\expandafter\ifx\csname natexlab\endcsname\relax\def\natexlab#1{#1}\fi

\bibitem[{Aharoni et~al.(2019)Aharoni, Johnson, and Firat}]{aharoni-etal-2019-massively}
Roee Aharoni, Melvin Johnson, and Orhan Firat. 2019.
\newblock \href {https://doi.org/10.18653/v1/N19-1388} {Massively multilingual neural machine translation}.
\newblock In \emph{Proceedings of the 2019 Conference of the North {A}merican Chapter of the Association for Computational Linguistics: Human Language Technologies, Volume 1 (Long and Short Papers)}, pages 3874--3884, Minneapolis, Minnesota. Association for Computational Linguistics.

\bibitem[{Al-Shedivat and Parikh(2019)}]{al2019consistency}
Maruan Al-Shedivat and Ankur~P Parikh. 2019.
\newblock \href {https://arxiv.org/pdf/1904.02338.pdf} {Consistency by agreement in zero-shot neural machine translation}.
\newblock \emph{arXiv preprint arXiv:1904.02338}.

\bibitem[{Arivazhagan et~al.(2019)Arivazhagan, Bapna, Firat, Aharoni, Johnson, and Macherey}]{arivazhagan2019missing}
Naveen Arivazhagan, Ankur Bapna, Orhan Firat, Roee Aharoni, Melvin Johnson, and Wolfgang Macherey. 2019.
\newblock \href {https://arxiv.org/pdf/1903.07091.pdf} {The missing ingredient in zero-shot neural machine translation}.
\newblock \emph{arXiv preprint arXiv:1903.07091}.

\bibitem[{Bahdanau et~al.(2014)Bahdanau, Cho, and Bengio}]{bahdanau2014neural}
Dzmitry Bahdanau, Kyunghyun Cho, and Yoshua Bengio. 2014.
\newblock \href {https://arxiv.org/pdf/1409.0473.pdf} {Neural machine translation by jointly learning to align and translate}.
\newblock \emph{arXiv preprint arXiv:1409.0473}.

\bibitem[{Ballesteros and Sanderson(2003)}]{ballesteros2003addressing}
Lisa Ballesteros and Mark Sanderson. 2003.
\newblock \href {https://dl.acm.org/doi/10.1145/956863.956891} {Addressing the lack of direct translation resources for cross-language retrieval}.
\newblock In \emph{Proceedings of the twelfth international conference on Information and knowledge management}, pages 147--152.

\bibitem[{Bapna et~al.(2022)Bapna, Caswell, Kreutzer, Firat, van Esch, Siddhant, Niu, Baljekar, Garcia, Macherey et~al.}]{bapna2022building}
Ankur Bapna, Isaac Caswell, Julia Kreutzer, Orhan Firat, Daan van Esch, Aditya Siddhant, Mengmeng Niu, Pallavi Baljekar, Xavier Garcia, Wolfgang Macherey, et~al. 2022.
\newblock \href {https://arxiv.org/pdf/2205.03983.pdf} {Building machine translation systems for the next thousand languages}.
\newblock \emph{arXiv preprint arXiv:2205.03983}.

\bibitem[{Bender et~al.(2021)Bender, Gebru, McMillan-Major, and Shmitchell}]{10.1145/3442188.3445922}
Emily~M. Bender, Timnit Gebru, Angelina McMillan-Major, and Shmargaret Shmitchell. 2021.
\newblock \href {https://doi.org/10.1145/3442188.3445922} {On the dangers of stochastic parrots: Can language models be too big?}
\newblock In \emph{Proceedings of the 2021 ACM Conference on Fairness, Accountability, and Transparency}, FAccT '21, page 610–623, New York, NY, USA. Association for Computing Machinery.

\bibitem[{Besacier(2014)}]{besacier-2014-machine}
Laurent Besacier. 2014.
\newblock \href {https://aclanthology.org/F14-2001} {Machine translation for litterature: a pilot study (traduction automatis{\'e}e d{'}une oeuvre litt{\'e}raire: une {\'e}tude pilote) [in {F}rench]}.
\newblock In \emph{Proceedings of TALN 2014 (Volume 2: Short Papers)}, pages 389--394, Marseille, France. Association pour le Traitement Automatique des Langues.

\bibitem[{Blodgett et~al.(2020)Blodgett, Barocas, Daum{\'e}~III, and Wallach}]{blodgett-etal-2020-language}
Su~Lin Blodgett, Solon Barocas, Hal Daum{\'e}~III, and Hanna Wallach. 2020.
\newblock \href {https://doi.org/10.18653/v1/2020.acl-main.485} {Language (technology) is power: A critical survey of {``}bias{''} in {NLP}}.
\newblock In \emph{Proceedings of the 58th Annual Meeting of the Association for Computational Linguistics}, pages 5454--5476, Online. Association for Computational Linguistics.

\bibitem[{Brown et~al.(2020)Brown, Mann, Ryder, Subbiah, Kaplan, Dhariwal, Neelakantan, Shyam, Sastry, Askell et~al.}]{brown2020language}
Tom Brown, Benjamin Mann, Nick Ryder, Melanie Subbiah, Jared~D Kaplan, Prafulla Dhariwal, Arvind Neelakantan, Pranav Shyam, Girish Sastry, Amanda Askell, et~al. 2020.
\newblock \href {https://proceedings.neurips.cc/paper/2020/hash/1457c0d6bfcb4967418bfb8ac142f64a-Abstract.html} {Language models are few-shot learners}.
\newblock \emph{Advances in neural information processing systems}, 33:1877--1901.

\bibitem[{Chen et~al.(2017)Chen, Liu, Cheng, and Li}]{chen2017teacher}
Yun Chen, Yang Liu, Yong Cheng, and Victor~OK Li. 2017.
\newblock \href {https://arxiv.org/pdf/1705.00753.pdf} {A teacher-student framework for zero-resource neural machine translation}.
\newblock \emph{arXiv preprint arXiv:1705.00753}.

\bibitem[{Chen et~al.(2018)Chen, Liu, and Li}]{chen2018zero}
Yun Chen, Yang Liu, and Victor Li. 2018.
\newblock \href {https://ojs.aaai.org/index.php/AAAI/article/view/11976} {Zero-resource neural machine translation with multi-agent communication game}.
\newblock In \emph{Proceedings of the aaai conference on artificial intelligence}, volume~32.

\bibitem[{Cheng and Cheng(2019)}]{cheng2019joint}
Yong Cheng and Yong Cheng. 2019.
\newblock \href {https://www.ijcai.org/proceedings/2017/0555.pdf} {Joint training for pivot-based neural machine translation}.
\newblock \emph{Joint training for neural machine translation}, pages 41--54.

\bibitem[{Chowdhery et~al.(2022)Chowdhery, Narang, Devlin, Bosma, Mishra, Roberts, Barham, Chung, Sutton, Gehrmann et~al.}]{chowdhery2022palm}
Aakanksha Chowdhery, Sharan Narang, Jacob Devlin, Maarten Bosma, Gaurav Mishra, Adam Roberts, Paul Barham, Hyung~Won Chung, Charles Sutton, Sebastian Gehrmann, et~al. 2022.
\newblock \href {https://arxiv.org/abs/2204.02311} {Palm: Scaling language modeling with pathways}.
\newblock \emph{arXiv preprint arXiv:2204.02311}.

\bibitem[{Currey and Heafield(2019)}]{currey2019zero}
Anna Currey and Kenneth Heafield. 2019.
\newblock \href {https://aclanthology.org/D19-5610.pdf} {Zero-resource neural machine translation with monolingual pivot data}.
\newblock In \emph{Proceedings of the 3rd Workshop on Neural Generation and Translation}, pages 99--107.

\bibitem[{Du et~al.(2022)Du, Huang, Dai, Tong, Lepikhin, Xu, Krikun, Zhou, Yu, Firat et~al.}]{du2022glam}
Nan Du, Yanping Huang, Andrew~M Dai, Simon Tong, Dmitry Lepikhin, Yuanzhong Xu, Maxim Krikun, Yanqi Zhou, Adams~Wei Yu, Orhan Firat, et~al. 2022.
\newblock \href {https://proceedings.mlr.press/v162/du22c.html} {Glam: Efficient scaling of language models with mixture-of-experts}.
\newblock In \emph{International Conference on Machine Learning}, pages 5547--5569. PMLR.

\bibitem[{ElNokrashy et~al.(2022)ElNokrashy, Hendy, Maher, Afify, and Awadalla}]{elnokrashy2022language}
Muhammad ElNokrashy, Amr Hendy, Mohamed Maher, Mohamed Afify, and Hany~Hassan Awadalla. 2022.
\newblock \href {https://arxiv.org/pdf/2208.05852.pdf} {Language tokens: A frustratingly simple approach improves zero-shot performance of multilingual translation}.
\newblock \emph{arXiv preprint arXiv:2208.05852}.

\bibitem[{Fedus et~al.(2022)Fedus, Dean, and Zoph}]{fedus2022review}
William Fedus, Jeff Dean, and Barret Zoph. 2022.
\newblock \href {https://arxiv.org/pdf/2209.01667.pdf} {A review of sparse expert models in deep learning}.
\newblock \emph{arXiv preprint arXiv:2209.01667}.

\bibitem[{Fedus et~al.(2021)Fedus, Zoph, and Shazeer}]{fedus2021switch}
William Fedus, Barret Zoph, and Noam Shazeer. 2021.
\newblock \href {https://www.jmlr.org/papers/volume23/21-0998/21-0998.pdf} {Switch transformers: Scaling to trillion parameter models with simple and efficient sparsity}.
\newblock \emph{J. Mach. Learn. Res}, 23:1--40.

\bibitem[{Firat et~al.(2016{\natexlab{a}})Firat, Cho, and Bengio}]{firat2016multi}
Orhan Firat, Kyunghyun Cho, and Yoshua Bengio. 2016{\natexlab{a}}.
\newblock \href {https://arxiv.org/pdf/1601.01073.pdf} {Multi-way, multilingual neural machine translation with a shared attention mechanism}.
\newblock \emph{arXiv preprint arXiv:1601.01073}.

\bibitem[{Firat et~al.(2016{\natexlab{b}})Firat, Sankaran, Al-Onaizan, Vural, and Cho}]{firat2016zero}
Orhan Firat, Baskaran Sankaran, Yaser Al-Onaizan, Fatos T~Yarman Vural, and Kyunghyun Cho. 2016{\natexlab{b}}.
\newblock \href {https://arxiv.org/pdf/1606.04164.pdf} {Zero-resource translation with multi-lingual neural machine translation}.
\newblock \emph{arXiv preprint arXiv:1606.04164}.

\bibitem[{Freitag and Firat(2020)}]{freitag2020complete}
Markus Freitag and Orhan Firat. 2020.
\newblock \href {https://arxiv.org/pdf/2010.10239.pdf} {Complete multilingual neural machine translation}.
\newblock \emph{arXiv preprint arXiv:2010.10239}.

\bibitem[{Freitag et~al.(2022)Freitag, Rei, Mathur, Lo, Stewart, Avramidis, Kocmi, Foster, Lavie, and Martins}]{freitag-etal-2022-results}
Markus Freitag, Ricardo Rei, Nitika Mathur, Chi-kiu Lo, Craig Stewart, Eleftherios Avramidis, Tom Kocmi, George Foster, Alon Lavie, and Andr{\'e} F.~T. Martins. 2022.
\newblock \href {https://aclanthology.org/2022.wmt-1.2} {Results of {WMT}22 metrics shared task: Stop using {BLEU} {--} neural metrics are better and more robust}.
\newblock In \emph{Proceedings of the Seventh Conference on Machine Translation (WMT)}, pages 46--68, Abu Dhabi, United Arab Emirates (Hybrid). Association for Computational Linguistics.

\bibitem[{Gambier and Van~Doorslaer(2016)}]{gambier2016border}
Yves Gambier and Luc Van~Doorslaer. 2016.
\newblock \emph{Border crossings: Translation studies and other disciplines}, volume 126.
\newblock John Benjamins Publishing Company.

\bibitem[{Gu et~al.(2019)Gu, Wang, Cho, and Li}]{gu2019improved}
Jiatao Gu, Yong Wang, Kyunghyun Cho, and Victor~OK Li. 2019.
\newblock \href {https://arxiv.org/pdf/1906.01181.pdf} {Improved zero-shot neural machine translation via ignoring spurious correlations}.
\newblock \emph{arXiv preprint arXiv:1906.01181}.

\bibitem[{Hassan et~al.(2018)Hassan, Aue, Chen, Chowdhary, Clark, Federmann, Huang, Junczys-Dowmunt, Lewis, Li et~al.}]{hassan2018achieving}
Hany Hassan, Anthony Aue, Chang Chen, Vishal Chowdhary, Jonathan Clark, Christian Federmann, Xuedong Huang, Marcin Junczys-Dowmunt, William Lewis, Mu~Li, et~al. 2018.
\newblock \href {https://arxiv.org/pdf/1803.05567.pdf} {Achieving human parity on automatic chinese to english news translation}.
\newblock \emph{arXiv preprint arXiv:1803.05567}.

\bibitem[{Hwang et~al.(2022)Hwang, Cui, Xiong, Yang, Liu, Hu, Wang, Salas, Jose, Ram et~al.}]{hwang2022tutel}
Changho Hwang, Wei Cui, Yifan Xiong, Ziyue Yang, Ze~Liu, Han Hu, Zilong Wang, Rafael Salas, Jithin Jose, Prabhat Ram, et~al. 2022.
\newblock \href {https://arxiv.org/pdf/2206.03382.pdf} {Tutel: Adaptive mixture-of-experts at scale}.
\newblock \emph{arXiv preprint arXiv:2206.03382}.

\bibitem[{Ji et~al.(2020)Ji, Zhang, Duan, Zhang, Chen, and Luo}]{ji2020cross}
Baijun Ji, Zhirui Zhang, Xiangyu Duan, Min Zhang, Boxing Chen, and Weihua Luo. 2020.
\newblock \href {https://ojs.aaai.org/index.php/AAAI/article/download/5341/5197} {Cross-lingual pre-training based transfer for zero-shot neural machine translation}.
\newblock In \emph{Proceedings of the AAAI conference on artificial intelligence}, volume~34, pages 115--122.

\bibitem[{Johnson et~al.(2017)Johnson, Schuster, Le, Krikun, Wu, Chen, Thorat, Vi{\'e}gas, Wattenberg, Corrado et~al.}]{johnson2017google}
Melvin Johnson, Mike Schuster, Quoc~V Le, Maxim Krikun, Yonghui Wu, Zhifeng Chen, Nikhil Thorat, Fernanda Vi{\'e}gas, Martin Wattenberg, Greg Corrado, et~al. 2017.
\newblock \href {https://aclanthology.org/Q17-1024.pdf} {Google’s multilingual neural machine translation system: Enabling zero-shot translation}.
\newblock \emph{Transactions of the Association for Computational Linguistics}, 5:339--351.

\bibitem[{Kim et~al.(2019)Kim, Petrov, Petrushkov, Khadivi, and Ney}]{kim2019pivot}
Yunsu Kim, Petre Petrov, Pavel Petrushkov, Shahram Khadivi, and Hermann Ney. 2019.
\newblock \href {https://arxiv.org/pdf/1909.09524.pdf} {Pivot-based transfer learning for neural machine translation between non-english languages}.
\newblock \emph{arXiv preprint arXiv:1909.09524}.

\bibitem[{Kudo and Richardson(2018)}]{kudo2018sentencepiece}
Taku Kudo and John Richardson. 2018.
\newblock \href {https://arxiv.org/pdf/1808.06226.pdf} {Sentencepiece: A simple and language independent subword tokenizer and detokenizer for neural text processing}.
\newblock \emph{arXiv preprint arXiv:1808.06226}.

\bibitem[{Kudugunta et~al.(2021)Kudugunta, Huang, Bapna, Krikun, Lepikhin, Luong, and Firat}]{kudugunta2021beyond}
Sneha Kudugunta, Yanping Huang, Ankur Bapna, Maxim Krikun, Dmitry Lepikhin, Minh-Thang Luong, and Orhan Firat. 2021.
\newblock \href {https://arxiv.org/pdf/2110.03742.pdf} {Beyond distillation: Task-level mixture-of-experts for efficient inference}.
\newblock \emph{arXiv preprint arXiv:2110.03742}.

\bibitem[{Lakew et~al.(2021)Lakew, Negri, and Turchi}]{lakew2021zero}
Surafel~M Lakew, Matteo Negri, and Marco Turchi. 2021.
\newblock \href {https://aclanthology.org/2021.mtsummit-loresmt.10.pdf} {Zero-shot neural machine translation with self-learning cycle}.
\newblock In \emph{Proceedings of the 4th Workshop on Technologies for MT of Low Resource Languages (LoResMT2021)}, pages 96--113.

\bibitem[{Lee-Thorp and Ainslie(2022)}]{lee2022sparse}
James Lee-Thorp and Joshua Ainslie. 2022.
\newblock \href {https://aclanthology.org/2022.findings-emnlp.5.pdf} {Sparse mixers: Combining moe and mixing to build a more efficient bert}.
\newblock \emph{arXiv preprint arXiv:2205.12399}.

\bibitem[{Lepikhin et~al.(2020)Lepikhin, Lee, Xu, Chen, Firat, Huang, Krikun, Shazeer, and Chen}]{lepikhin2020gshard}
Dmitry Lepikhin, HyoukJoong Lee, Yuanzhong Xu, Dehao Chen, Orhan Firat, Yanping Huang, Maxim Krikun, Noam Shazeer, and Zhifeng Chen. 2020.
\newblock \href {https://arxiv.org/pdf/2006.16668.pdf} {Gshard: Scaling giant models with conditional computation and automatic sharding}.
\newblock \emph{arXiv preprint arXiv:2006.16668}.

\bibitem[{Liu et~al.(2020)Liu, Niehues, Cross, Guzm{\'a}n, and Li}]{liu2020improving}
Danni Liu, Jan Niehues, James Cross, Francisco Guzm{\'a}n, and Xian Li. 2020.
\newblock \href {https://aclanthology.org/2021.acl-long.101.pdf} {Improving zero-shot translation by disentangling positional information}.
\newblock \emph{arXiv preprint arXiv:2012.15127}.

\bibitem[{Lu et~al.(2018)Lu, Keung, Ladhak, Bhardwaj, Zhang, and Sun}]{lu2018neural}
Yichao Lu, Phillip Keung, Faisal Ladhak, Vikas Bhardwaj, Shaonan Zhang, and Jason Sun. 2018.
\newblock \href {https://arxiv.org/pdf/1804.08198.pdf} {A neural interlingua for multilingual machine translation}.
\newblock \emph{arXiv preprint arXiv:1804.08198}.

\bibitem[{Mustafa et~al.(2022)Mustafa, Riquelme, Puigcerver, Jenatton, and Houlsby}]{https://doi.org/10.48550/arxiv.2206.02770}
Basil Mustafa, Carlos Riquelme, Joan Puigcerver, Rodolphe Jenatton, and Neil Houlsby. 2022.
\newblock \href {https://doi.org/10.48550/ARXIV.2206.02770} {Multimodal contrastive learning with limoe: the language-image mixture of experts}.

\bibitem[{Papineni et~al.(2002)Papineni, Roukos, Ward, and Zhu}]{papineni-etal-2002-bleu}
Kishore Papineni, Salim Roukos, Todd Ward, and Wei-Jing Zhu. 2002.
\newblock \href {https://doi.org/10.3115/1073083.1073135} {{B}leu: a method for automatic evaluation of machine translation}.
\newblock In \emph{Proceedings of the 40th Annual Meeting of the Association for Computational Linguistics}, pages 311--318, Philadelphia, Pennsylvania, USA. Association for Computational Linguistics.

\bibitem[{Philip et~al.(2020)Philip, Berard, Gall{\'e}, and Besacier}]{philip-etal-2020-monolingual}
Jerin Philip, Alexandre Berard, Matthias Gall{\'e}, and Laurent Besacier. 2020.
\newblock \href {https://doi.org/10.18653/v1/2020.emnlp-main.361} {Monolingual adapters for zero-shot neural machine translation}.
\newblock In \emph{Proceedings of the 2020 Conference on Empirical Methods in Natural Language Processing (EMNLP)}, pages 4465--4470, Online. Association for Computational Linguistics.

\bibitem[{Puigcerver et~al.(2020)Puigcerver, Riquelme, Mustafa, Renggli, Pinto, Gelly, Keysers, and Houlsby}]{DBLP:journals/corr/abs-2009-13239}
Joan Puigcerver, Carlos Riquelme, Basil Mustafa, C{\'{e}}dric Renggli, Andr{\'{e}}~Susano Pinto, Sylvain Gelly, Daniel Keysers, and Neil Houlsby. 2020.
\newblock \href {http://arxiv.org/abs/2009.13239} {Scalable transfer learning with expert models}.
\newblock \emph{CoRR}, abs/2009.13239.

\bibitem[{Rei et~al.(2022)Rei, C.~de Souza, Alves, Zerva, Farinha, Glushkova, Lavie, Coheur, and Martins}]{rei-etal-2022-comet}
Ricardo Rei, Jos{\'e}~G. C.~de Souza, Duarte Alves, Chrysoula Zerva, Ana~C Farinha, Taisiya Glushkova, Alon Lavie, Luisa Coheur, and Andr{\'e} F.~T. Martins. 2022.
\newblock \href {https://aclanthology.org/2022.wmt-1.52} {{COMET}-22: Unbabel-{IST} 2022 submission for the metrics shared task}.
\newblock In \emph{Proceedings of the Seventh Conference on Machine Translation (WMT)}, pages 578--585, Abu Dhabi, United Arab Emirates (Hybrid). Association for Computational Linguistics.

\bibitem[{Rios et~al.(2020)Rios, M{\"u}ller, and Sennrich}]{rios2020subword}
Annette Rios, Mathias M{\"u}ller, and Rico Sennrich. 2020.
\newblock \href {https://arxiv.org/pdf/2011.01703.pdf} {Subword segmentation and a single bridge language affect zero-shot neural machine translation}.
\newblock \emph{arXiv preprint arXiv:2011.01703}.

\bibitem[{Riquelme et~al.(2021)Riquelme, Puigcerver, Mustafa, Neumann, Jenatton, Susano~Pinto, Keysers, and Houlsby}]{riquelme2021scaling}
Carlos Riquelme, Joan Puigcerver, Basil Mustafa, Maxim Neumann, Rodolphe Jenatton, Andr{\'e} Susano~Pinto, Daniel Keysers, and Neil Houlsby. 2021.
\newblock \href {https://proceedings.neurips.cc/paper/2021/file/48237d9f2dea8c74c2a72126cf63d933-Paper.pdf} {Scaling vision with sparse mixture of experts}.
\newblock \emph{Advances in Neural Information Processing Systems}, 34:8583--8595.

\bibitem[{Ryabinin and Gusev(2020)}]{ryabinin2020crowdsourced}
Max Ryabinin and Anton Gusev. 2020.
\newblock \href {http://arxiv.org/abs/2002.04013} {Towards crowdsourced training of large neural networks using decentralized mixture-of-experts}.

\bibitem[{Sellam et~al.(2020)Sellam, Das, and Parikh}]{sellam-etal-2020-bleurt}
Thibault Sellam, Dipanjan Das, and Ankur Parikh. 2020.
\newblock \href {https://doi.org/10.18653/v1/2020.acl-main.704} {{BLEURT}: Learning robust metrics for text generation}.
\newblock In \emph{Proceedings of the 58th Annual Meeting of the Association for Computational Linguistics}, pages 7881--7892, Online. Association for Computational Linguistics.

\bibitem[{Shazeer et~al.(2017)Shazeer, Mirhoseini, Maziarz, Davis, Le, Hinton, and Dean}]{shazeer2017outrageously}
Noam Shazeer, Azalia Mirhoseini, Krzysztof Maziarz, Andy Davis, Quoc Le, Geoffrey Hinton, and Jeff Dean. 2017.
\newblock \href {https://arxiv.org/pdf/1701.06538.pdf} {Outrageously large neural networks: The sparsely-gated mixture-of-experts layer}.
\newblock \emph{arXiv preprint arXiv:1701.06538}.

\bibitem[{Sutskever et~al.(2014)Sutskever, Vinyals, and Le}]{sutskever2014sequence}
Ilya Sutskever, Oriol Vinyals, and Quoc~V Le. 2014.
\newblock \href {https://proceedings.neurips.cc/paper/2014/file/a14ac55a4f27472c5d894ec1c3c743d2-Paper.pdf} {Sequence to sequence learning with neural networks}.
\newblock \emph{Advances in neural information processing systems}, 27.

\bibitem[{Taivalkoski-Shilov(2019)}]{taivalkoski2019ethical}
Kristiina Taivalkoski-Shilov. 2019.
\newblock \href {https://www.tandfonline.com/doi/pdf/10.1080/0907676X.2018.1520907?needAccess=true&role=button} {Ethical issues regarding machine (-assisted) translation of literary texts}.
\newblock \emph{Perspectives}, 27(5):689--703.

\bibitem[{Tang et~al.(2021)Tang, Tran, Li, Chen, Goyal, Chaudhary, Gu, and Fan}]{tang-etal-2021-multilingual}
Yuqing Tang, Chau Tran, Xian Li, Peng-Jen Chen, Naman Goyal, Vishrav Chaudhary, Jiatao Gu, and Angela Fan. 2021.
\newblock \href {https://doi.org/10.18653/v1/2021.findings-acl.304} {Multilingual translation from denoising pre-training}.
\newblock In \emph{Findings of the Association for Computational Linguistics: ACL-IJCNLP 2021}, pages 3450--3466, Online. Association for Computational Linguistics.

\bibitem[{Tiedemann(2018)}]{tiedemann2018emerging}
J{\"o}rg Tiedemann. 2018.
\newblock \href {https://arxiv.org/pdf/1802.00273.pdf} {Emerging language spaces learned from massively multilingual corpora}.
\newblock \emph{arXiv preprint arXiv:1802.00273}.

\bibitem[{Vaswani et~al.(2017)Vaswani, Shazeer, Parmar, Uszkoreit, Jones, Gomez, Kaiser, and Polosukhin}]{vaswani2017attention}
Ashish Vaswani, Noam Shazeer, Niki Parmar, Jakob Uszkoreit, Llion Jones, Aidan~N Gomez, {\L}ukasz Kaiser, and Illia Polosukhin. 2017.
\newblock \href {https://arxiv.org/pdf/1706.03762.pdf} {Attention is all you need}.
\newblock \emph{Advances in neural information processing systems}, 30.

\bibitem[{Wang et~al.(2021)Wang, Zhang, Du, Chen, Xie, and Luo}]{wang-etal-2021-rethinking-zero}
Weizhi Wang, Zhirui Zhang, Yichao Du, Boxing Chen, Jun Xie, and Weihua Luo. 2021.
\newblock \href {https://doi.org/10.18653/v1/2021.findings-emnlp.366} {Rethinking zero-shot neural machine translation: From a perspective of latent variables}.
\newblock In \emph{Findings of the Association for Computational Linguistics: EMNLP 2021}, pages 4321--4327, Punta Cana, Dominican Republic. Association for Computational Linguistics.

\bibitem[{Wu et~al.(2022)Wu, Liu, Chen, Chen, Dai, and Yuan}]{wu2022residual}
Lemeng Wu, Mengchen Liu, Yinpeng Chen, Dongdong Chen, Xiyang Dai, and Lu~Yuan. 2022.
\newblock \href {https://arxiv.org/pdf/2204.09636.pdf} {Residual mixture of experts}.
\newblock \emph{arXiv preprint arXiv:2204.09636}.

\bibitem[{Xu et~al.(2022)Xu, Yang, Meng et~al.}]{xu2022eag}
Yulin Xu, Zhen Yang, Fandong Meng, et~al. 2022.
\newblock \href {https://arxiv.org/pdf/2203.02180.pdf} {Eag: Extract and generate multi-way aligned corpus for complete multi-lingual neural machine translation}.
\newblock \emph{arXiv preprint arXiv:2203.02180}.

\bibitem[{Yang et~al.(2022)Yang, Yin, Ma, Zhang, Wu, Guo, Li, and Wei}]{Yang_2022}
Jian Yang, Yuwei Yin, Shuming Ma, Dongdong Zhang, Shuangzhi Wu, Hongcheng Guo, Zhoujun Li, and Furu Wei. 2022.
\newblock \href {https://doi.org/10.24963/ijcai.2022/618} {{UM}4: Unified multilingual multiple teacher-student model for zero-resource neural machine translation}.
\newblock In \emph{Proceedings of the Thirty-First International Joint Conference on Artificial Intelligence}. International Joint Conferences on Artificial Intelligence Organization.

\bibitem[{Yang et~al.(2021)Yang, Eriguchi, Muzio, Tadepalli, Lee, and Hassan}]{yang-etal-2021-improving-multilingual}
Yilin Yang, Akiko Eriguchi, Alexandre Muzio, Prasad Tadepalli, Stefan Lee, and Hany Hassan. 2021.
\newblock \href {https://doi.org/10.18653/v1/2021.emnlp-main.578} {Improving multilingual translation by representation and gradient regularization}.
\newblock In \emph{Proceedings of the 2021 Conference on Empirical Methods in Natural Language Processing}, pages 7266--7279, Online and Punta Cana, Dominican Republic. Association for Computational Linguistics.

\bibitem[{You et~al.(2021)You, Feng, Su, and Yu}]{DBLP:journals/corr/abs-2105-03036}
Zhao You, Shulin Feng, Dan Su, and Dong Yu. 2021.
\newblock \href {http://arxiv.org/abs/2105.03036} {Speechmoe: Scaling to large acoustic models with dynamic routing mixture of experts}.
\newblock \emph{CoRR}, abs/2105.03036.

\bibitem[{You et~al.(2022)You, Feng, Su, and Yu}]{you2022speechmoe2}
Zhao You, Shulin Feng, Dan Su, and Dong Yu. 2022.
\newblock \href {https://ieeexplore.ieee.org/stamp/stamp.jsp?arnumber=9747065} {Speechmoe2: Mixture-of-experts model with improved routing}.
\newblock In \emph{ICASSP 2022-2022 IEEE International Conference on Acoustics, Speech and Signal Processing (ICASSP)}, pages 7217--7221. IEEE.

\bibitem[{Zhang et~al.(2020)Zhang, Williams, Titov, and Sennrich}]{zhang2020improving}
Biao Zhang, Philip Williams, Ivan Titov, and Rico Sennrich. 2020.
\newblock \href {https://arxiv.org/pdf/2004.11867.pdf} {Improving massively multilingual neural machine translation and zero-shot translation}.
\newblock \emph{arXiv preprint arXiv:2004.11867}.

\bibitem[{Zheng et~al.(2017)Zheng, Cheng, and Liu}]{zheng2017maximum}
Hao Zheng, Yong Cheng, and Yang Liu. 2017.
\newblock \href {https://www.ijcai.org/proceedings/2017/0594.pdf} {Maximum expected likelihood estimation for zero-resource neural machine translation.}
\newblock In \emph{IJCAI}, pages 4251--4257.

\bibitem[{Zoph et~al.(2022)Zoph, Bello, Kumar, Du, Huang, Dean, Shazeer, and Fedus}]{zoph2022designing}
Barret Zoph, Irwan Bello, Sameer Kumar, Nan Du, Yanping Huang, Jeff Dean, Noam Shazeer, and William Fedus. 2022.
\newblock \href {https://arxiv.org/abs/2202.08906} {St-moe: Designing stable and transferable sparse expert models}.
\newblock \emph{arXiv preprint arXiv:2202.08906}.

\bibitem[{Zoph and Knight(2016)}]{zoph-knight-2016-multi}
Barret Zoph and Kevin Knight. 2016.
\newblock \href {https://doi.org/10.18653/v1/N16-1004} {Multi-source neural translation}.
\newblock In \emph{Proceedings of the 2016 Conference of the North {A}merican Chapter of the Association for Computational Linguistics: Human Language Technologies}, pages 30--34, San Diego, California. Association for Computational Linguistics.

\end{thebibliography}
\bibliographystyle{acl_natbib}
\newpage \clearpage
\appendix
\section*{Appendix}
\label{appendix}
\begin{figure*}
\centering
\includegraphics[scale=0.12]{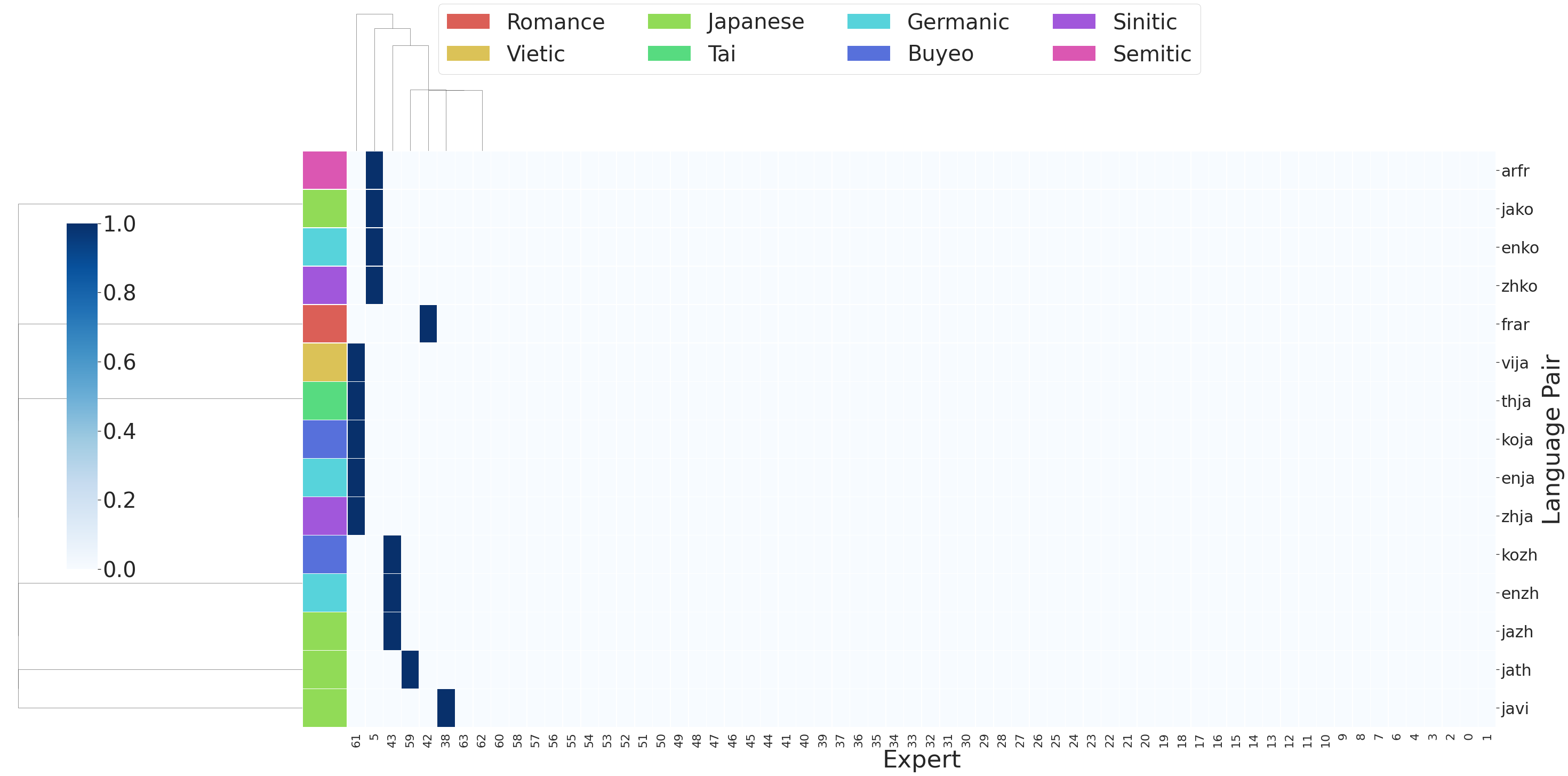} 
\caption{Routing decisions of the last encoder layer of our Task-level MoE model with 64 experts, trained with pair target language to task id mapping, with tl\_a used during inference, for 2M steps.}
\label{fig:fig_64_tl_enc3_1m}
\end{figure*}

\begin{figure*}
\vspace{-20mm}
\centering
\includegraphics[scale=0.12]{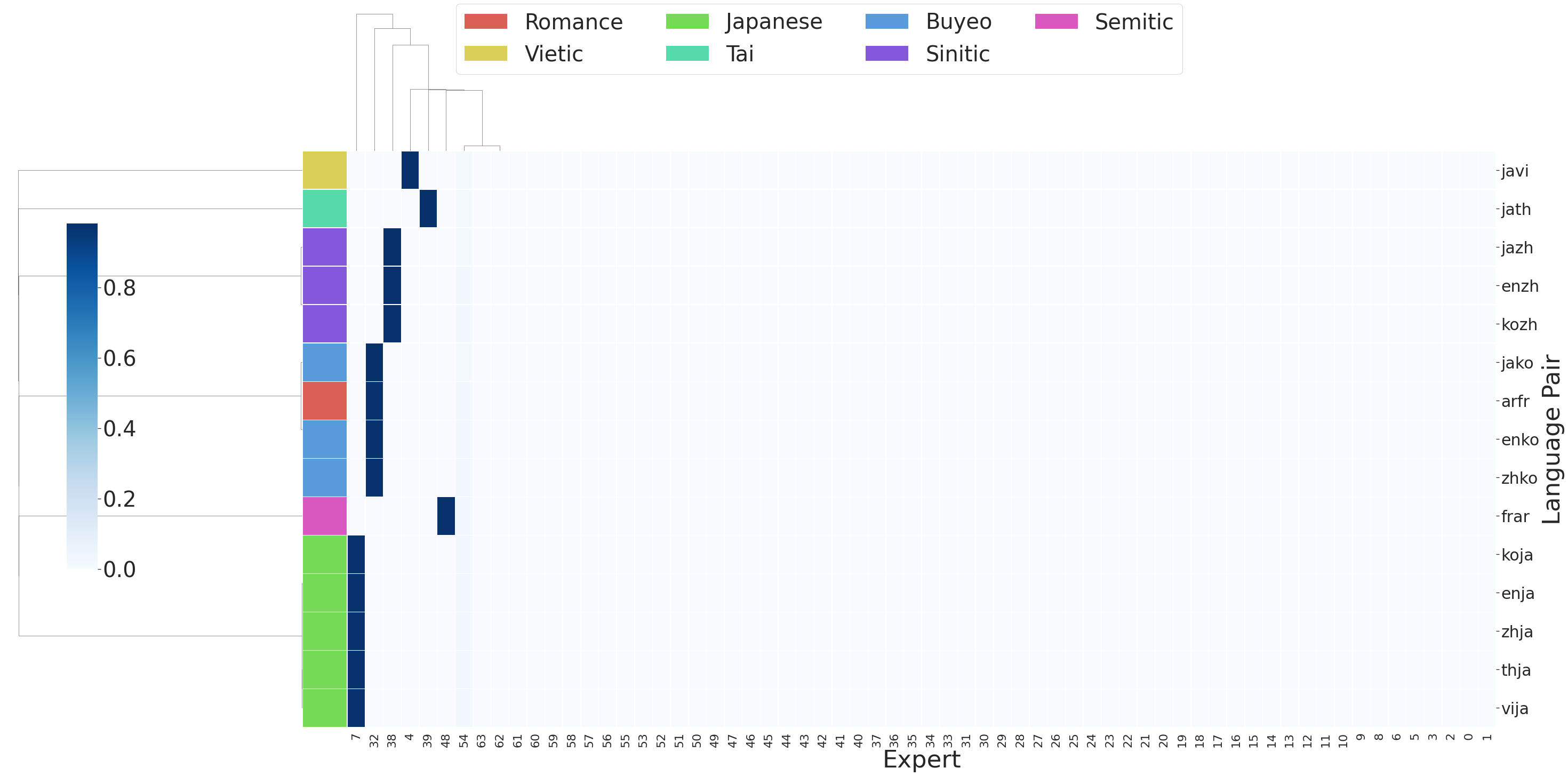}
\caption{Routing decisions of the last decoder layer of our Task-level MoE model with 64 experts, trained with pair target language to task id mapping, with tl\_a used during inference, for 2M steps.}
\label{fig:fig_64_tl_dec3_1m}
\end{figure*}

In Figures \ref{fig:fig_64_tl_enc3_1m}, \ref{fig:fig_64_tl_dec3_1m} we see the routing decisions maps for the last encoder and decoder layer of TL- based task-level MoE models trained with 64 experts and tl\_a mapping during inference. The distribution to the experts is similarly expected here, following the training and inference pairs-to-experts' mapping, as same target language pairs get routed to the same experts. The overlap in the encoder and decoder experts is minimal. All but one selected experts are different between the encoder and decoder.

\vspace{10mm}
\newpage

\begin{table*}[!htbp]
\begin{tabular}{ll|ll|ll}
\toprule
\multicolumn{1}{l}{\textbf{LP}} & \multicolumn{1}{l}{\textbf{Num. of sentences}}|&\multicolumn{1}{l}{\textbf{LP}} & \multicolumn{1}{l}{\textbf{Num. of sentences}}|&\multicolumn{1}{l}{\textbf{LP}} & \multicolumn{1}{l}{\textbf{Num. of sentences}} \\ \hline
ar--fr                                        & 0                                                 &
fr--ar                                        & 0                                                 &
be--ru                                        & 3,641                                              \\
ru--be                                        & 3,641                                              &
ja--th                                        & 6,200                                              &
th--ja                                        & 6,200                                              \\
ko--th                                        & 6,208                                              &
th--zh                                        & 12,305                                             &
zh--th                                        & 12,305                                             \\
id--ms                                        & 24,535                                             &
kk--tr                                        & 69,476                                             &
tr--kk                                        & 69,476                                             \\
ug--en                                        & 129,688                                            &
tk--en                                        & 140,247                                            &
en--tk                                        & 144,094                                            \\
en--ug                                        & 167,741                                            &
kk--ru                                        & 177,931                                            &
ru--kk                                        & 177,931                                            \\
bg--mk                                        & 205,651                                            &
mk--bg                                        & 205,651                                            &
or--en                                        & 262,099                                            \\
cs--es                                        & 285,546                                            &
es--cs                                        & 285,546                                            &
ko--vi                                        & 369,594                                            \\
vi--ko                                        & 369,594                                            &
et--fi                                        & 466,254                                            &
fi--et                                        & 466,254                                            \\
ko--zh                                        & 498,968                                            &
zh--ko                                        & 498,968                                            &
vi--zh                                        & 504,037                                            \\
zh--vi                                        & 504,037                                            &
en--tt                                        & 556,751                                            &
de--ru                                        & 585,161                                            \\
ru--de                                        & 585,161                                            &
ja--vi                                        & 604,940                                            &
vi--ja                                        & 604,940                                            \\
zh--tr                                        & 609,066                                            &
en--or                                        & 658,015                                            &
en--rw                                        & 792,559                                            \\
tt--en                                        & 819,579                                            &
cs--fr                                        & 895,912                                            &
fr--cs                                        & 895,912                                            \\
ja--ko                                        & 974,896                                            &
ko--ja                                        & 974,896                                            &
cs--de                                        & 1,007,448                                           \\
de--cs                                        & 1,007,448                                           &
rw--en                                        & 1,264,813                                           &
ja--zh                                        & 1,339,622                                           \\
zh--ja                                        & 1,339,622                                           &
ru--tr                                        & 1,467,160                                           &
tr--ru                                        & 1,467,160                                           \\
cs--ru                                        & 1,644,795                                           &
ru--cs                                        & 1,644,795                                           &
en--la                                        & 1,809,003                                           \\
en--xh                                        & 1,895,974                                           &
la--en                                        & 2,167,039                                           &
en--sm                                        & 2,353,608                                           \\
de--es                                        & 2,886,854                                           &
de--fr                                        & 2,886,854                                           &
es--de                                        & 2,886,854                                           \\
en--st                                        & 2,931,225                                           &
en--sn                                        & 3,022,722                                           &
en--ig                                        & 3,058,540                                           \\
en--haw                                       & 3,393,044                                           &
en--yo                                        & 3,404,254                                           &
en--lo                                        & 3,573,485                                           \\
fr--de                                        & 3,681,694                                           &
ru--uk                                        & 3,792,978                                           &
uk--ru                                        & 3,792,978                                           \\
en--yi                                        & 3,904,022                                           &
en--ku                                        & 5,254,260                                           &
en--ny                                        & 5,408,239                                           \\
en--so                                        & 5,472,253                                           &
xh--en                                        & 5,728,449                                           &
sm--en                                        & 5,730,016                                           \\
lo--en                                        & 5,766,691                                           &
en--mi                                        & 5,793,624                                           &
en--co                                        & 6,847,178                                           \\
ig--en                                        & 7,224,372                                           &
en--tg                                        & 8,136,182                                           &
sn--en                                        & 8,319,539                                           \\
yi--en                                        & 8,347,482                                           &
en--zu                                        & 8,376,755                                           &
haw--en                                       & 8,646,782                                           \\
en--ha                                        & 8,995,196                                           &
en--sd                                        & 9,095,582                                           &
en--ps                                        & 9,446,484                                           \\
yo--en                                        & 9,565,981                                           &
st--en                                        & 10,052,460                                          &
en--ky                                        & 10,213,875                                          \\
mi--en                                        & 10,383,802                                          &
en--ht                                        & 11,758,532                                          &
ku--en                                        & 11,984,479                                          \\
en--ga                                        & 12,494,184                                          &
ny--en                                        & 12,616,421                                          &
so--en                                        & 12,745,144                                          \\
en--su                                        & 14,088,043                                          &
en--am                                        & 14,583,643                                          &
en--mt                                        & 15,693,771                                          \\
en--pa                                        & 15,784,955                                          &
en--ceb                                       & 15,965,215                                          &
ga--en                                        & 16,145,438                                          \\
en--mg                                        & 17,474,049                                          &
co--en                                        & 17,490,449                                          &
en--be                                        & 17,645,900                                          \\
en--eo                                        & 18,104,746                                          &
es--ru                                        & 18,363,087                                          &
ru--es                                        & 18,363,087                                          \\
tg--en                                        & 18,509,169                                          &
en--fy                                        & 19,495,193                                          &
ha--en                                        & 19,687,379                                          \\
en--mn                                        & 20,012,644                                          &
sd--en                                        & 20,337,227                                          &
es--fr                                        & 20,792,781                                          \\
fr--es                                        & 20,792,781                                          &
zu--en                                        & 21,010,498                                          &
ky--en                                        & 21,357,639                                          \\
ps--en                                        & 21,613,753                                          &
ht--en                                        & 22,455,117                                          &
en--gd                                        & 22,495,036                                          \\
fr--ru                                        & 22,890,960                                          &
ru--fr                                        & 22,890,960                                          &
en--eu                                        & 23,279,562                                          \\
en--my                                        & 23,848,386                                          &
en--hy                                        & 25,779,512                                          &
en--lb                                        & 25,943,545                                          \\
be--en                                        & 28,585,822                                          &
en--hmn                                       & 29,258,771                                          &
pa--en                                        & 29,592,573                                          \\
su--en                                        & 30,234,526                                          &
en--jv                                        & 30,315,732                                          &
en--uz                                        & 30,943,271                                          \\
en--si                                        & 31,847,809                                          &
en--kk                                        & 32,075,645                                          &
en--cy                                        & 32,089,017                                          \\
am--en                                        & 32,110,524                                          &
en--ml                                        & 32,335,094                                          &
mg--en                                        & 32,356,401                                          \\
en--gu                                        & 32,742,711                                          &
fy--en                                        & 33,389,347                                          &
en--mk                                        & 33,546,951                                          \\
en--mr                                        & 33,629,504                                          &
hy--en                                        & 34,146,553                                          &
eu--en                                        & 35,158,779                                          \\
hmn--en                                       & 35,795,085                                          &
en--bs                                        & 35,801,135                                          &
gd--en                                        & 36,082,889                                          \\
en--kn                                        & 36,472,587                                          &
en--km                                        & 37,229,149                                          &
en--hr                                        & 37,248,537                                          \\
en--ne                                        & 38,606,074                                          &
en--sw                                        & 41,198,888                                          &
mt--en                                        & 41,723,852             
\\ \bottomrule
\end{tabular}
\end{table*}

\begin{table*}[!htbp]
\begin{tabular}{ll|ll|ll}
\toprule
\multicolumn{1}{l}{\textbf{LP}} & \multicolumn{1}{l}{\textbf{Num. of sentences}}|&\multicolumn{1}{l}{\textbf{LP}} & \multicolumn{1}{l}{\textbf{Num. of sentences}}|&\multicolumn{1}{l}{\textbf{LP}} & \multicolumn{1}{l}{\textbf{Num. of sentences}} \\ \hline
en--ka                                        & 42,992,730                                          &
en--te                                        & 43,112,561                                          &
en--gl                                        & 45,053,738                                          \\
lb--en                                        & 46,042,055                                          &
kn--en                                        & 46,847,179                                          &
en--sr                                        & 48,786,193                                          \\
mn--en                                        & 49,510,283                                          &
cy--en                                        & 50,252,795                                          &
en--af                                        & 50,525,196                                          \\
gu--en                                        & 53,217,408                                          &
en--sq                                        & 55,374,368                                          &
my--en                                        & 55,886,059                                          \\
jv--en                                        & 57,400,218                                          &
ceb--en                                       & 59,055,133                                          &
mk--en                                        & 61,867,654                                          \\
en--ta                                        & 62,352,671                                          &
eo--en                                        & 62,552,455                                          &
en--az                                        & 64,274,455                                          \\
kk--en                                        & 64,724,987                                          &
ne--en                                        & 69,288,148                                          &
si--en                                        & 69,667,030                                          \\
en--lv                                        & 69,998,174                                          &
ml--en                                        & 70,059,552                                          &
en--is                                        & 71,837,680                                          \\
hr--en                                        & 72,301,373                                          &
en--ur                                        & 74,947,168                                          &
en--et                                        & 75,581,372                                          \\
te--en                                        & 75,967,139                                          &
en--bn                                        & 80,362,347                                          &
km--en                                        & 80,514,558                                          \\
en--uk                                        & 86,997,037                                          &
en--ca                                        & 87,267,893                                          &
en--sk                                        & 89,304,972                                          \\
gl--en                                        & 89,695,691                                          &
bs--en                                        & 90,263,002                                          &
sw--en                                        & 90,493,834                                          \\
af--en                                        & 94,492,215                                          &
en--fil                                       & 97,193,343                                          &
en--sl                                        & 97,924,827                                          \\
ka--en                                        & 99,344,373                                          &
ta--en                                        & 100,659,244                                         &
en--lt                                        & 102,081,855                                         \\
is--en                                        & 103,425,627                                         &
sq--en                                        & 104,573,595                                         &
mr--en                                        & 105,247,834                                         \\
uz--en                                        & 105,701,543                                         &
az--en                                        & 115,304,502                                         &
en--iw                                        & 117,411,687                                         \\
en--ms                                        & 117,562,320                                         &
en--fa                                        & 117,720,531                                         &
en--bg                                        & 128,893,255                                         \\
en--fi                                        & 133,307,056                                         &
en--el                                        & 135,547,629                                         &
en--ro                                        & 136,723,562                                         \\
ca--en                                        & 137,014,332                                         &
sr--en                                        & 137,913,360                                         &
en--hu                                        & 146,210,128                                         \\
ur--en                                        & 149,852,215                                         &
fil--en                                       & 156,986,036                                         &
lv--en                                        & 163,533,602                                         \\
en--no                                        & 165,068,512                                         &
en--cs                                        & 167,802,683                                         &
en--da                                        & 176,676,171                                         \\
sl--en                                        & 189,776,212                                         &
bn--en                                        & 190,650,787                                         &
et--en                                        & 197,134,423                                         \\
uk--en                                        & 209,888,151                                         &
en--sv                                        & 217,128,777                                         &
lt--en                                        & 245,092,644                                         \\
bg--en                                        & 289,282,624                                         &
sk--en                                        & 295,623,418                                         &
fi--en                                        & 300,151,301                                         \\
el--en                                        & 322,429,692                                         &
th--en                                        & 340,378,951                                         &
ro--en                                        & 352,986,741                                         \\
en--th                                        & 353,001,311                                         &
fa--en                                        & 402,836,484                                         &
ms--en                                        & 414,693,796                                         \\
iw--en                                        & 427,057,571                                         &
hu--en                                        & 436,082,952                                         &
hi--en                                        & 490,464,288                                         \\
en--hi                                        & 494,638,634                                         &
ar--en                                        & 525,411,304                                         &
id--en                                        & 527,345,119                                         \\
en--ar                                        & 532,778,336                                         &
ko--en                                        & 535,863,357                                         &
en--id                                        & 538,581,029                                         \\
da--en                                        & 539,573,001                                         &
en--ko                                        & 546,191,335                                         &
no--en                                        & 613,221,604                                         \\
vi--en                                        & 632,740,475                                         &
en--vi                                        & 656,115,977                                         &
cs--en                                        & 661,833,636                                         \\
sv--en                                        & 800,882,060                                         &
ja--en                                        & 846,991,020                                         &
tr--en                                        & 850,154,583                                         \\
en--tr                                        & 869,295,018                                         &
en--ja                                        & 876,842,917                                         &
it--en                                        & 998,195,505                                         \\
en--it                                        & 100,981,5294                                        &
pl--en                                        & 111,935,3071                                        &
en--pl                                        & 1,141,598,628                                        \\
pt--en                                        & 1,155,038,272                                        &
en--pt                                        & 1,184,401,180                                        &
en--zh                                        & 1,228,817,744                                        \\
zh--en                                        & 1,238,691,743                                        &
ru--en                                        & 1,425,268,039                                        &
en--ru                                        & 1,455,103,126                                        \\
nl--en                                        & 1,580,819,532                                        &
en--nl                                        & 1,587,530,791                                        &
de--en                                        & 1,680,270,443                                        \\
en--de                                        & 1,695,095,726                                        &
en--fr                                        & 1,887,609,530                                        &
fr--en                                        & 1,922,463,803                                        \\
en--es                                        & 2,419,825,975                                        &
es--en                                        & 2,435,228,645                                        \\
\bottomrule
\end{tabular}
\caption{Number of Sentences of each Language Pair (LP) in our train set.}
\label{table:train_num_of_sents}
\end{table*}

\end{document}